%% file: main.tex
\documentclass[nohyperref]{article}

\usepackage{microtype}
\usepackage{graphicx}
\usepackage{subfigure}
\usepackage{booktabs} % for professional tables
\usepackage{amsfonts}
\usepackage{xcolor}
\usepackage{amsmath}
\usepackage{enumitem}
\usepackage{multirow}

\definecolor{highergreen}{RGB}{26,152,80} 
\definecolor{bestblue}{RGB}{215,48,39}  % red
\definecolor{worsered}{RGB}{33,102,172}  % blue

\DeclareMathOperator*{\argmin}{arg\,min}

\usepackage{hyperref}

\usepackage{amsmath}
\usepackage{amssymb}
\usepackage{mathtools}
\usepackage{amsthm}

\usepackage[accepted]{icml2022}

\icmltitlerunning{RankSim:  Ranking Similarity Regularization for Deep Imbalanced Regression}

\begin{document}

\twocolumn[
\icmltitle{RankSim:  Ranking Similarity Regularization for Deep Imbalanced Regression}



\icmlsetsymbol{equal}{*}

\begin{icmlauthorlist}

\icmlauthor{Yu Gong}{equal,yyy,comp}
\icmlauthor{Greg Mori}{yyy,comp}
\icmlauthor{Frederick Tung}{comp}

\end{icmlauthorlist}

\icmlaffiliation{yyy}{Simon Fraser University}
\icmlaffiliation{comp}{Borealis AI}

\icmlcorrespondingauthor{Yu Gong}{gongyug@sfu.ca}

\icmlkeywords{Machine Learning, ICML}

\vskip 0.3in
]



\printAffiliationsAndNotice{\icmlEqualContribution} % otherwise use the standard text.

\begin{abstract}
Data imbalance, in which a plurality of the data samples come from a small proportion of labels, poses a challenge in training deep neural networks. 
Unlike classification, in regression the labels are continuous, potentially boundless, and form a natural ordering. These distinct features of regression call for new techniques that leverage the additional information encoded in label-space relationships. This paper presents the RankSim (ranking similarity) regularizer for deep imbalanced regression, which encodes an inductive bias that samples that are closer in label space should also be closer in feature space. In contrast to recent distribution smoothing based approaches, RankSim captures both nearby and distant relationships: for a given data sample, RankSim encourages the sorted list of its neighbors in label space to match the sorted list of its neighbors in feature space. RankSim is complementary to conventional imbalanced learning techniques, including re-weighting, two-stage training, and distribution smoothing, and lifts the state-of-the-art performance on three imbalanced regression benchmarks: IMDB-WIKI-DIR, AgeDB-DIR, and STS-B-DIR.
\end{abstract}

\input{chapters/introduction}
\input{chapters/related_work}

\input{chapters/method}
\input{chapters/experiments}

\input{chapters/conclusion}

\bibliography{reference}
\bibliographystyle{icml2022}

\onecolumn
\newpage
\appendix
\input{chapters/supp}

\end{document}

%% file: chapters/introduction.tex
\section{Introduction}

Skewed data distributions, in which a plurality of the data instances are from a small number of labels, widely exist in the real world and pose challenges to conventional machine learning techniques. Although deep learning has made much progress and shown its potential in many application domains, imbalanced datasets often present practical difficulties, causing models to exhibit undesirable bias towards the majority labels and under-representing the minority labels.
Prior work has thoroughly studied how to improve model training on imbalanced data, such as via cost-sensitive learning, representation learning, and decoupled learning. However, previous attempts have almost exclusively focused on the task of imbalanced classification.  A very recent work~\cite{yangetal2021} drew attention to the nuances of regression with imbalanced data and identified the task of deep imbalanced regression as a distinct problem. 

Compared to classification, how to train deep regression models with imbalanced data is not yet as well understood.
Unlike classification networks predicting discrete labels to model the categorical distributions, regression networks aim to predict labels with continuous values. The continuity in label space makes deep imbalanced regression different from deep imbalanced classification.
On the one hand, the target values can be infinite and boundless, which makes many methods designed for deep imbalanced classification untenable. On the other hand, the continuity in label space can also play a positive role by providing extra information about data instance relationships.

\begin{figure}[t]
\label{fig:illustration}
\includegraphics[width=\linewidth]{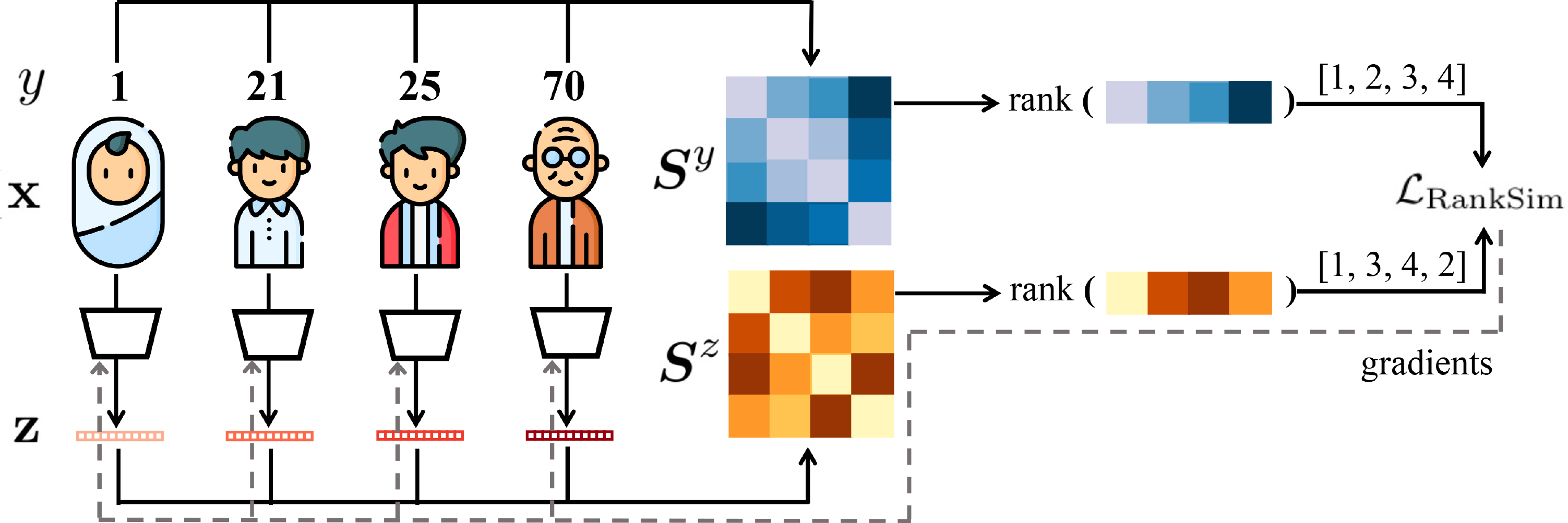}
\caption{The RankSim (ranking similarity) regularizer for deep imbalanced regression introduces an inductive bias that items that are closer in label space should also be closer in feature space. Matrices $\boldsymbol{S}^y$ and $\boldsymbol{S}^z$ encode pairwise similarities in label space and feature space, respectively. For a given input sample, RankSim encourages the sorted list of its neighbors in label space to match the sorted list of its neighbors in feature space. }
\centering
\end{figure}

In regression, the prediction targets form a natural ordering. For example, if we are interested in age estimation, we may order people from youngest to oldest; if we are performing home valuation, we may order properties from cheapest to most expensive. We can use this natural ordering of labels to regularize the representation learned by a neural network. For example, we can nudge the network in a direction such that the feature representation for the face of a 21-year old is more similar to the representation for a 25-year old than for a 70-year-old. This intuition of preserving label-space relationships in the learned feature space is the motivation behind recent smoothing-based regularization approaches to address the imbalanced data problem in regression, such as label distribution smoothing (LDS) and feature distribution smoothing (FDS) \cite{yangetal2021}.
As an illustrative example of why preserving label-space relationships helps with imbalanced regression, when \citet{yangetal2021} trained vanilla models on an imbalanced age estimation dataset, they observed that the learned features of 0-6 year-olds, which have very few samples, were highly similar to the learned features of 30 year-olds, which have a large number of samples. This proximity in feature space, despite the large difference in label space, is undesirable as it hinders generalization to unseen 0-6 year-olds.

Our proposed RankSim regularizer builds on this intuition and applies a stronger inductive bias. We provide an illustraton in Fig.~\ref{fig:illustration}. Using age estimation as an example, suppose we have images of individuals whose ages are $1, 21, 25$, and $70$. Denote, with abuse of superscripting notation for simplicity of illustration, their learned feature representations by $\mathbf{z}^1, \mathbf{z}^{21}, \mathbf{z}^{25}, \mathbf{z}^{70}$. Smoothing-based regularization approaches would encourage, for example, $\mathbf{z}^{21}$ to be similar to $\mathbf{z}^{25}$. However, we might also like $\mathbf{z}^{21}$ to be somewhat similar to $\mathbf{z}^{1}$, yet not as similar as $\mathbf{z}^{21}$ is to $\mathbf{z}^{25}$. Let $\sigma(\cdot, \cdot)$ denote a similarity function over vectors, such as the cosine similarity. Restating our preference above, we would like $\sigma(\mathbf{z}^{21}, \mathbf{z}^{25}) > \sigma(\mathbf{z}^{21}, \mathbf{z}^{1})$. Extending the idea, we can construct a complete desired ordering based on the distance in label space (age) from the anchor $\mathbf{z}^{21}$: $\sigma(\mathbf{z}^{21}, \mathbf{z}^{25}) > \sigma(\mathbf{z}^{21}, \mathbf{z}^{1}) > \sigma(\mathbf{z}^{21}, \mathbf{z}^{70})$. This construction can be repeated for the other samples as anchors. For example, using $\mathbf{z}^{25}$ as anchor: $\sigma(\mathbf{z}^{25}, \mathbf{z}^{21}) >  \sigma(\mathbf{z}^{25}, \mathbf{z}^{1}) > \sigma(\mathbf{z}^{25}, \mathbf{z}^{70})$. In general, these conditions encode an inductive bias that, for a given input sample, items that are closer in label space are also closer in feature space.

The rest of this paper is structured as follows. We next provide an overview of related work in deep imbalanced classification and regression. We then describe the RankSim regularizer for deep imbalanced regression, which encodes a more ``global" inductive bias on label-space/feature-space relationships than state-of-the-art smoothing approaches: RankSim's inductive bias captures not only nearby relationships but also distant relationships in label and feature space. Finally, we present new state-of-the-art results on three public benchmarks for deep imbalanced regression, as well as ablation experiments and analysis. {Code and pretrained weights are available at \url{https://github.com/BorealisAI/ranksim-imbalanced-regression}.}

%% file: chapters/related_work.tex
\section{Related Work}
\paragraph{Imbalanced Classification.}
Prior work on learning with imbalanced data has predominantly focused on the imbalanced (or long-tailed) classification problem, where many classes have very few instances. Approaches can be grouped into data-based and model-based paradigms. 

\textit{Data-based} methods are associated with the input data. One strategy is re-sampling---over-sampling the minority classes~\cite{smote2002, Byrd2019WhatIT} or under-sampling the majority classes~\cite{Han2005BorderlineSMOTEAN, Buda2018ASS}. 
Recent work~\cite{chang2021image} introduced the re-sampling strategy to imbalanced object detection with a novel object-centric memory replay framework.
Another option is data augmentation. Implicit semantic data augmentation (ISDA)~\cite{NEURIPS2019_15f99f21} estimates the covariance matrices of each class to obtain semantic directions and generate augmented samples;
Remix~\cite{10.1007/978-3-030-65414-6_9} and UniMix~\cite{xu2021towards} extend mixup~\cite{zhang2018mixup} to enhance minority classes.
\textit{Model-based} methods address the imbalance problem from a model perspective. 
Logit adjustment~\cite{tian2020posterior, tang2020longtailed} relies on post-hoc correction to adjust the decision boundary after training.
Cost-sensitive learning~\cite{lintsungyi2018, NEURIPS2019_621461af, Ren2020balms} aims at adjusting losses for different classes during training, considering that the costs caused by different classes are not equal. For example, focal loss~\cite{lintsungyi2018} inversely re-weights the loss with the prediction probabilities based on the observation that the minority classes suffer from achieving low loss values.
Representation learning approaches~\cite{yang2020rethinking, kang2021exploring, jiang2021self} primarily help the model focus on learning a less biased feature extractor.
Two-stage (or decoupled) training, introduced by~\citet{Kang2020Decoupling}, splits the learning procedure into separate stages for representation learning and classifier learning.
A recent work~\cite{zhong2021mislas} introduced label-aware smoothing to handle different degrees of over-confidence for classes and improve classifier learning in the second stage.

% \vspace{-3mm}

\paragraph{Imbalanced Regression.} 

The continuity in label space makes imbalanced regression different from imbalanced classification. On the one hand, the target values can be infinite and boundless, which makes many methods designed for deep imbalanced classification untenable. On the other hand, the continuity in label space can provide extra information about data instance relationships.

Early work~\cite{smote2002,torgoetal2013,brancoetal2017} on imbalanced regression attempted to re-sample the training set by synthesizing new samples for minority targets. 
To build meaningful connections between different labels and handle potential missing values, state-of-the-art approaches proposed to use kernel density estimation to perform smoothing on the target distribution. Label distribution smoothing (LDS)~\cite{yangetal2021} and DenseLoss~\cite{michael2021} share the similar idea of applying a Gaussian kernel to the empirical label density to estimate an ``effective'' label density distribution that takes the continuity of labels into account. 
Feature distribution smoothing (FDS)~\cite{yangetal2021}, aiming to introduce continuity of the feature, performs distribution smoothing on the feature space by transferring the feature statistics (mean and covariance) between nearby targets. However, these methods only account for the nearby label values, i.e. they encode a ``local" inductive bias. 
In contrast, RankSim captures both nearby and distant relationships in label and feature space, encoding a more ``global" inductive bias.

Some methods for imbalanced classification can also be adapted to regression. For example, inspired by Focal loss~\cite{lintsungyi2018}, a regression variant Focal-R~\cite{yangetal2021} re-weights the loss values by the continuous $L_1$ error. We will show in the experiments that RankSim is complementary to Focal-R and other conventional imbalanced learning techniques.  Progressive Margin Loss (PML)~\cite{pml2021} focuses on imbalanced age estimation. It constructs the age relationship by learning label distributions, and handles the imbalanced data with margin-based imbalanced classification techniques. 
% ~\citet{}
% \vspace{-mm}
\paragraph{Ranking-based Methods.}
The optimization of ranking-based losses is challenging due to the non-differentiability of the ranking operation.
To approximate the non-differentiable and non-decomposable ranking operation,~\citet{taylor08} considered the expectation of the ranking under a random perturbation; 
\citet{song16icml} leveraged the dynamic programming algorithm to update the weights; 
\citet{He_2018_TALR},~\citet{cakiretal2019} and~\citet{revaud19iccv} optimized a differentiable relaxation of histogram binning~\cite{NIPS2016_325995af};
\citet{engilberge2019sodeep} proposed to learn an LSTM-based surrogate network; 
\citet{NEURIPS2019_d8c24ca8} formulated it as an optimal transport problem. 
In our implementation, we recast the ranking operation as the minimizer of a linear combinatorial objective~\cite{rolineketal2020} and leverage an elegant method for efficient backpropagation through blackbox combinatorial solvers~\cite{vlastelicaetal2020} (details in Section 3). 
While prior work focused on directly optimizing ranking-based metrics such as Average Precision (AP) or Normalized Discounted Cumulative Gain (NDCG), we aim to make the ranking of feature-space neighbors to be similar to the ranking of label-space neighbors.

%% file: chapters/method.tex
\section{Method}
\label{sec:method}

To formalize the proposed regularizer, we first set up some preliminaries. Let $\mathbf{a} \in \mathbb{R}^n$ be an arbitrary vector of $n$ real values. Let $\mathbf{rk}$ denote the ranking function, such that $\mathbf{rk}(\mathbf{a})$ is the permutation of $\{1, ..., n\}$ containing the rank (in sorted order) of each element in $\mathbf{a}$. That is, the $i$th element in $\mathbf{rk}(\mathbf{a})$ is given by
\begin{equation}
\mathbf{rk}(\mathbf{a})_i = 1 + |\{j: \mathbf{a}_j > \mathbf{a}_i\}|.
\end{equation}

\noindent For example, if $\mathbf{a} = [9, 5, 11, 6]$, then $\mathbf{rk}(\mathbf{a}) = [2, 4, 1, 3]$. 
$\mathbf{rk}(\mathbf{a})_1$ is 2 because $\mathbf{a}_1$ is the second-largest (rank = 2) element in $\mathbf{a}$; $\mathbf{rk}(\mathbf{a})_2$ is 4 because $\mathbf{a}_2$ is the fourth-largest element in $\mathbf{a}$;  $\mathbf{rk}(\mathbf{a})_3$ is 1 because $\mathbf{a}_3$ is the largest element in $\mathbf{a}$; finally, $\mathbf{rk}(\mathbf{a})_4$ is 3 because $\mathbf{a}_4$ is the third-largest element in $\mathbf{a}$.

Let our regression dataset consist of a set of pairs $(\mathbf{x}_i, y_i)$, where $\mathbf{x}_i$ denotes the input and $y_i$ the corresponding continuous-value target\footnote{In the following manuscript, the \textit{target} is used interchangeably with the \textit{label}.}. Denote by $\mathbf{z} = f(\mathbf{x}; \theta)$ the feature representation of $\mathbf{x}$ generated by a neural network parameterized by $\theta$.
The idea behind RankSim (ranking similarity) regularization is to encourage  alignment between the ranking of neighbors in label space (the $y$'s) and the ranking of neighbors in feature space (the $\mathbf{z}$'s).

Consider a subset of pairs $\mathcal{M} = \{(\mathbf{x}_i, y_i), i=1, ..., M\}$. Let $\boldsymbol{S}^y \in \mathbb{R}^{M \times M}$ be the pairwise similarity matrix obtained by applying similarity function $\sigma^y$ in label space across all elements in $\mathcal{M}$. In other words, the $(i, j)$th entry in $\boldsymbol{S}^y$ is given by
\begin{equation}
\boldsymbol{S}^y_{i,j} = \sigma^y(y_i, y_j),
\end{equation}
which is the similarity between items $i$ and $j$ in label space. For continuous scalar labels, $\sigma^y$ can simply return the negative absolute distance. Analogously, let $\boldsymbol{S}^{z} \in \mathbb{R}^{M \times M}$ be the pairwise similarity matrix obtained by applying similarity function $\sigma^{z}$ in feature space. The $(i,j)$th entry in $\boldsymbol{S}^z$ is given by
\begin{equation}\label{eq:feat_sim_matrix}
\boldsymbol{S}^{z}_{i,j} = \sigma^{z}(\mathbf{z}_i, \mathbf{z}_j) = \sigma^{z}(f(\mathbf{x}_i; \theta), f(\mathbf{x}_j; \theta)),
\end{equation}
which is the similarity between items $i$ and $j$ in feature space. $\sigma^z$ is a similarity function defined over vectors, such as the cosine similarity. 

We can now define the RankSim regularization loss with respect to subset $\mathcal{M}$:
\begin{equation}
\label{eq:RankSim}
\mathcal{L}_{\operatorname{RankSim}} = \sum_{i =1}^{|\mathcal{M}|} \ell \, (\mathbf{rk}( \boldsymbol{S}^y_{[i,:]}), \, \mathbf{rk}(\boldsymbol{S}^{z}_{[i,:]})),
\end{equation}

\noindent where $[i,:]$ denotes the $i$th row in the matrix and ranking similarity function $\ell$ penalizes differences in the input vectors. Concretely, we adopt the mean squared error for $\ell$, which makes minimizing $\ell$ equivalent to maximizing the Spearman correlation of the label-space and feature-space ranking vectors.
During training, $\mathcal{M}$ is constructed from each batch; to reduce ties and boost the relative representation of infrequent labels, we sample $\mathcal{M}$ from the current batch such that each label occurs at most once.

Intuitively, Eq.~\ref{eq:RankSim} says that, given an input $(\mathbf{x}_i, y_i)$, we would like the sorted list of its neighbors in label space to match the sorted list of its neighbors in feature space as closely as possible. When the match is exact, the loss is zero. Using the previous age estimation example, since $\sigma^y(y^{11}, y^9) > \sigma^y(y^{11}, y^6) > \sigma^y(y^{11}, y^5) > \sigma^y(y^{11}, y^{55}) > \sigma^y(y^{11}, y^{60})$, where $\sigma^y$ is the negative absolute distance, we would like $\sigma^z(\mathbf{z}^{11}, \mathbf{z}^9) > \sigma^z(\mathbf{z}^{11}, \mathbf{z}^6) > \sigma^z(\mathbf{z}^{11}, \mathbf{z}^5) > \sigma^z(\mathbf{z}^{11}, \mathbf{z}^{55}) > \sigma^z(\mathbf{z}^{11}, \mathbf{z}^{60})$, where $\sigma^z$ is the cosine similarity, for example.

 Eq.~\ref{eq:RankSim} is challenging to optimize because of the non-differentiability of the ranking operation. Ranking-based losses are piecewise constant functions of their input, and the gradient is zero almost everywhere: intuitively, small changes in the input to the ranking function do not always result in a change in the output ranking. However, we can recast the ranking operation as the minimizer of a linear combinatorial objective \cite{rolineketal2020}:
\begin{equation}
\mathbf{rk}(\mathbf{a}) = \argmin_{\pi \in \Pi_n} \mathbf{a} \cdot \pi,
\end{equation}
where $\Pi_n$ is the set of all permutations of $\{1, ..., n\}$. This allows us leverage an elegant method for efficient backpropagation through blackbox combinatorial solvers \cite{vlastelicaetal2020}.
To obtain an informative gradient from the piecewise constant loss landscape, \cite{vlastelicaetal2020} implicitly constructs a family of piecewise affine continuous interpolation functions parameterized by a single hyperparameter $\lambda > 0$ that trades off the informativeness of the gradient with fidelity to the original function. During the backward pass, instead of returning the true gradient (zero almost everywhere), we compute and return the gradient of the continuous interpolation:

\begin{equation}
\label{eq:dlda}
    \frac{\partial \mathcal{L}}{\partial \mathbf{a}} = -\frac{1}{\lambda} (\mathbf{rk}(\mathbf{a}) - \mathbf{rk}(\mathbf{a}_\lambda)),
\end{equation}

\noindent where $\mathbf{a}_\lambda$ is constructed based on the incoming gradient information $\frac{\partial \mathcal{L}}{\partial \mathbf{rk}}$ by

\begin{equation}
\mathbf{a}_\lambda = \mathbf{a} + \lambda \, \frac{\partial \mathcal{L}}{\partial \mathbf{rk}} \, .
\end{equation}

We can therefore backpropagate through the ranking operations in Eq. \ref{eq:RankSim} at the cost of an additional call to $\mathbf{rk}$ (i.e., the call in Eq. \ref{eq:dlda} on the perturbed input). For clarity, the blackbox ``solver" in our case is simply the ranking function $\mathbf{rk}$, which can be implemented by sorting operations; there is no need for a general combinatorial solver.

The resulting regularizer is straightforward to implement, can be computed in closed form, and introduces only two hyperparameters: the interpolation strength $\lambda$, and the balancing weight $\gamma$ on the regularization term in the overall network loss (i.e., adding $\gamma \mathcal{L}_{\operatorname{RankSim}}$ to the other loss terms).

%% file: chapters/experiments.tex
\section{Experiments}

We performed extensive experimental validation of RankSim on three public benchmarks for deep imbalanced regression. We first describe the benchmarks, metrics, and baselines. We then present experimental results on the three benchmarks, including comparisons with state-of-the-art approaches. Finally, we present ablation studies and analysis to better understand the impact of our design choices. 

\paragraph{Benchmarks.} 
We consider three datasets recently introduced by \citet{yangetal2021}.
\textbf{\textit{IMDB-WIKI-DIR}} is an age estimation dataset derived from \textit{IMDB-WIKI}~\cite{Rothe2016DeepEO}, which consists of face images with age annotations. It has 191,509 training samples, 11,022 validation samples and 11,022 test samples;  
\textbf{\textit{AgeDB-DIR}} is an age estimation dataset derived from \textit{AgeDB}~\cite{Moschoglou2017AgeDBTF}, with 12,208 samples for training, 2,140 samples for validation, and 2,140 samples for testing;
\textbf{\textit{STS-B-DIR}} is a natural language dataset derived from the \textit{STS-B}~\cite{cer-etal-2017-semeval,wang-etal-2018-glue}, which provides (continuous) similarity score between sentence pairs.
It contains 5,249 training sentence pairs, 1,000 validation pairs and 1,000 test pairs. 
For further details on the benchmarks, including label distribution plots, we refer the interested reader to \citet{yangetal2021}. 
All datasets contain imbalanced training set and balanced validation/test sets.

\paragraph{Evaluation Metrics.}

Following \citet{yangetal2021} and common practice in imbalanced learning \cite{liuetal2019}, we report overall results on the whole test set, as well as on the subsets of \textit{many-shot region} (bins with $>$100 training samples),
\textit{medium-shot region} (bins with 20 to 100 training samples), and \textit{few-shot region} (bins with $<$20 training samples). In \textit{IMDB-WIKI-DIR} and \textit{AgeDB-DIR}, each bin is 1 year. In \textit{STS-B-DIR}, the bin size is 0.1.
We report mean absolute error (MAE, lower is better) and geometric mean (GM, lower is better) on \textit{IMDB-WIKI-DIR} and \textit{AgeDB-DIR}, and mean squared error (MSE, lower is better) and Pearson correlation (higher is better) on \textit{STS-B-DIR}. Additional metrics are provided in the supplementary.

\paragraph{Baselines.} 

For fair comparison, we adopt the standard network architectures specified in \citet{yangetal2021}, i.e., ResNet-50 for \textit{IMDB-WIKI-DIR} and \textit{AgeDB-DIR}, and BiLSTM + GloVe word embeddings for \textit{STS-B-DIR}.
RankSim is orthogonal to conventional imbalanced learning techniques, such as re-weighting~\cite{lintsungyi2018} and two-stage (or decoupled) training \cite{Kang2020Decoupling}, as well as to state-of-the-art distribution smoothing approaches for deep imbalanced regression \cite{yangetal2021}. We apply RankSim to regularize standard methods including Vanilla, Focal-R, RRT, SQINV (or INV), and apply the state-of-the-art LDS and FDS~\cite{yangetal2021}.
The full details about methods and training are described in the supplementary~\ref{supp:mode_baseline}. 
We present detailed experimental analysis verifying that, not only does RankSim lift the state-of-the-art performance on the three benchmarks, it is in fact \textit{complementary} to these well-established approaches.

\subsection{IMDB-WIKI-DIR}

\input{chapters/tables/table_imdb}

Table~\ref{table:imdb-wiki-dir} shows experimental results on the \textit{IMDB-WIKI-DIR} benchmark. The baseline numbers are quoted from \citet{yangetal2021}. The table is grouped into four sections: vanilla (base network), Focal-R (focal loss \cite{lintsungyi2018} adapted to regression), regressor retraining (two-stage training \cite{Kang2020Decoupling} adapted to regression, abbreviated RRT), and square-root inverse frequency re-weighting (SQINV). Within each group, we first show the results obtained by applying the state-of-the-art label and feature distribution smoothing methods (LDS and FDS, respectively) \cite{yangetal2021}, both separately and in combination. We integrated RankSim on top of each of these baselines.

Within each of the four sections, the best results are highlighted in bold. RankSim consistently obtains the best result within each section, across both metrics (MAE, GM) and test subsets (all, many-shot, medium-shot, few-shot). This outcome indicates that RankSim is complementary to standard imbalanced learning techniques.

For each metric and test subset, the best overall result is highlighted in red. RankSim sets a new state-of-the-art across all metrics and test subsets. For example, when combined with two-stage retraining (RRT) and feature distribution smoothing (FDS), RankSim achieves a leading mean absolute error of 7.35 on the overall test set. In the few-shot region, RankSim combined with re-weighting (SQINV), label distribution smoothing (LDS), and feature distribution smoothing (FDS) achieves a leading mean absolute error of 21.43. 

\subsection{AgeDB-DIR}

\input{chapters/tables/table_agedb}

Table~\ref{table:agedb} shows experimental results on the AgeDB-DIR benchmark, following the same structure as \textit{IMDB-WIKI-DIR} above. Comparing within each section (vanilla, Focal-R, RRT, SQINV), RankSim obtains the best result in 31 out of the 32 combinations of metric (MAE, GM) and test subset (all, many-shot, medium-shot, few-shot), again demonstrating that RankSim is complementary to standard imbalanced learning techniques. 
Overall, RankSim sets a new state-of-the-art on 7 out of 8 metric-subset combinations\footnote{In our experiments, we observed that a smaller batch size than recommended in \citet{yangetal2021} improved baseline performance. We include these improved baseline numbers in the supplementary. RankSim outperforms the baselines in both settings.}. For example, when combined with square-root inverse frequency re-weighting, RankSim achieves a leading mean absolute error of 6.91 on the overall test set. In the few-shot setting, RankSim with square-root inverse frequency re-weighting and feature distribution smoothing achieves a leading mean absolute error of 9.68.

\input{chapters/tables/table_sts_short}

\subsection{STS-B-DIR}

Table~\ref{table:sts-b1} shows experimental results on the STS-B-DIR benchmark. INV refers to inverse frequency re-weighting. RankSim consistently achieves state-of-the-art performance on the test set overall, as well as in the medium-shot and few-shot categories; the INV+LDS+FDS baseline performs slightly better in the many-shot category. The best overall results are obtained by RankSim combined with two-stage retraining (RRT): 0.865 mean squared error and 77.1\% Pearson correlation.
In the few-shot setting, RankSim combined with RRT achieves a leading 0.670 mean squared error and 86.1\% Pearson correlation.

\input{chapters/ablation}

%% file: chapters/tables/table_imdb.tex
\begin{table}[t]
% \ra{1.3}
\centering
\caption{\textbf{Results on \textit{IMDB-WIKI-DIR}}. Baseline numbers are quoted from \cite{yangetal2021}. The {best} results for each method (Vanilla, Focal-R, RRT, SQINV) are in \textbf{bold}. The best results for each metric and data subset (entire column) are in \textbf{\textcolor{bestblue}{bold and red}}.}
\label{table:imdb-wiki-dir}
\setlength{\tabcolsep}{2.5pt}
\resizebox{0.96\linewidth}{!}{\begin{tabular}{@{}lllllcllll@{}}\toprule[1.2pt]
& \multicolumn{4}{c}{\textbf{MAE $\downarrow$}} & \phantom{abc}& \multicolumn{4}{c}{\textbf{GM $\downarrow$}} \\
\cmidrule{2-5} \cmidrule{7-10} 
& All & Man. & Med. & Few &&  All & Man. & Med. & Few \\
\midrule[1.2pt]
\textsc{Vanilla} 
& 8.06 & 7.23 & 15.12 & 26.33 && 4.57 & 4.17 & 10.59 & 20.46 \\
  + \textbf{\textsc{RankSim}} 
  & 7.72 & \textbf{6.93} & 14.48 & 25.38 && {4.27} & \textbf{3.90} & 10.02 & {17.84}  \\
 + \textsc{LDS}  
 & 7.83 & 7.31 & {12.43} & 22.51 && 4.42 & 4.19 & {7.00} & 13.94 \\ 
 + \textsc{LDS} + \textbf{\textsc{RankSim}}  & \textbf{7.57} & {7.00} & \textbf{12.16} & {22.44} && \textbf{4.23} & {4.00} & {6.81} & 13.23 \\
 + \textsc{FDS}  
 &  7.85 & {7.18} & 13.35 & 24.12 && 4.47 & 4.18 & 8.18 & 15.18 \\ 
 + \textsc{FDS} + \textbf{\textsc{RankSim}} & {7.74} & \textbf{6.93} & {14.71} & {24.91} && {4.34} & {3.96} & {10.35} & {16.85} \\ 
 + \textsc{LDS} + \textsc{FDS} 
 & {7.78} & {7.20} & {12.61} & {22.19} && {4.37} & {4.12} & {7.39} & {12.61}\\
  + \textsc{LDS} + \textsc{FDS} + \textbf{\textsc{RankSim}} 
& {7.69} & {7.13} & {12.30} & \textcolor{bestblue}{\textbf{21.43}} && {4.34} & 4.13 & \textbf{6.72} &  \textcolor{bestblue}{\textbf{12.48}} \\
\midrule
\textsc{Focal-R}
& 7.97 & 7.12 & 15.14 & 26.96 && 4.49 & 4.10 & 10.37 & 21.20 \\ 
 + \textbf{\textsc{RankSim}}   
& {7.77} & {6.99} & 14.23 & 26.01 && {4.38} & {4.03} & 9.25 & 20.16  \\
 + \textsc{LDS}  
& 7.90 & 7.10 & {14.72} & 25.84 && 4.47 & {4.09} & {10.11} & 19.14\\ 
 + \textsc{LDS} + \textbf{\textsc{RankSim}}   
& 7.71 & 6.99 & \textbf{13.65} & {24.97} && 4.31 & 3.98 & \textbf{8.72} & {17.56} \\
 + \textsc{FDS}  
& 7.96 & 7.14 & {14.71} & 26.06 && 4.51 & 4.12 & {10.16} & 19.56\\ 
 + \textsc{FDS} + \textbf{\textsc{RankSim}} & {7.75} & 7.01 & {14.06} & \textbf{24.56} && {4.33} & {3.99} & {9.04} & \textbf{16.26} \\
 +  \textsc{LDS}   +  \textsc{FDS}   
& {7.88} & {7.10} & {14.08} & {25.75} && {4.47} & 4.11 & {9.32} & {18.67}\\
+ \textsc{LDS} + \textsc{FDS} + \textbf{\textsc{RankSim}} & \textbf{7.67} & \textbf{6.91} & 14.07 & 25.01 && \textbf{4.28} & \textbf{3.93} & 9.38 & 18.41 \\
\midrule
\textsc{RRT}      
& 7.81 & 7.07 & 14.06 & 25.13 && 4.35 & 4.03 & 8.91 & 16.96 \\ 
 + \textbf{\textsc{RankSim}}    
& 7.55 & 6.83 & 13.47 & 24.72 && 4.17 & {3.86} & 8.66 & 15.54\\
 + \textsc{LDS}           
& 7.79 & 7.08 & 13.76 & 24.64 && 4.34 & {4.02} & 8.72 & 16.92 \\ 
 + \textsc{LDS} + \textbf{\textsc{RankSim}}   
& 7.56 & 6.83 & 13.06 & 24.78 && 4.23 & 3.91 & 8.55 & 17.44 \\
 + \textsc{FDS}    
 & 7.65 &  {7.02} & 12.68 & 23.85 && 4.31 & 4.03 & 7.58 & 16.28 \\ 
  + \textsc{FDS} + \textbf{\textsc{RankSim}} 
&  \textcolor{bestblue}{\textbf{7.35}} & 6.81 & \textcolor{bestblue}{\textbf{11.50}} & \textbf{22.75} && \textcolor{bestblue}{\textbf{4.05}} & {3.85} & \textcolor{bestblue}{\textbf{6.05}} & \textbf{14.68}\\
 + \textsc{LDS} + \textsc{FDS}     
&  {7.65} & 7.06 &  {12.41} & {23.51} && {4.31} & 4.07 & {7.17} & {15.44} \\
 + \textsc{LDS} + \textsc{FDS} + \textbf{\textsc{RankSim}} &  {7.37} & \textcolor{bestblue}{\textbf{6.80}} &  {11.83} & {23.11} && 4.06 & \textcolor{bestblue}{\textbf{3.84}} & {6.33} & {14.71} \\ 
 \midrule
  \textsc{SQINV}  
  & 7.87 & 7.24 & 12.44 & 22.76 && 4.47 & 4.22 & 7.25 & 15.10 \\ 
   + \textbf{\textsc{RankSim}}   
   & \textcolor{black}{\textbf{7.42}} & \textbf{6.84} & \textbf{12.12} & 22.13 && \textbf{4.10} & \textbf{3.87} & {6.74} &  {12.78} \\
+ \textsc{LDS}  & 7.83 & 7.31 & 12.43 & 22.51 && 4.42 & 4.19 & 7.00 & 13.94 \\ 
 + \textsc{LDS} + \textbf{\textsc{RankSim}}  
& {7.57} & {7.00} & {12.16} & {22.44} && {4.23} & {4.00} & {6.81} & 13.23 \\
+ \textsc{FDS}           & 7.83 & 7.23 & 12.60 & 22.37 && 4.42 & 4.20 & 6.93 & 13.48\\  + \textsc{FDS}  + \textbf{\textsc{RankSim}} 
& 7.50 & {6.93} & 12.09 & 21.68 && {4.19} & {3.97} & \textbf{6.65} & 13.28\\
 + \textsc{LDS}   + \textsc{FDS}     & 7.78 & 7.20 & 12.61 & {22.19} && 4.37 & 4.12 & 7.39 &  {12.61} \\
 + \textsc{LDS} + \textsc{FDS}  + \textbf{\textsc{RankSim}}  
& {7.69} & {7.13} & {12.30} & \textcolor{bestblue}{\textbf{21.43}} && {4.34} & 4.13 & {6.72} &  \textcolor{bestblue}{\textbf{12.48}}\\ \midrule
\textbf{\textsc{Ours}} \textsc{vs.} \textbf{\textsc{Vanilla}} & \textbf{\textcolor{highergreen}{+0.71}} & \textbf{\textcolor{highergreen}{+0.43}} & \textbf{\textcolor{highergreen}{+3.62}} & \textbf{\textcolor{highergreen}{+4.90}} && \textbf{\textcolor{highergreen}{+0.52}} & \textbf{\textcolor{highergreen}{+0.33}} & \textbf{\textcolor{highergreen}{+4.54}} & \textbf{\textcolor{highergreen}{+7.98}} \\ 
\textbf{\textsc{Ours}} \textsc{vs.} \textbf{Yang et al. } & \textbf{\textcolor{highergreen}{+0.30}} & \textbf{\textcolor{highergreen}{+0.22}} & \textbf{\textcolor{highergreen}{+0.91}} & \textbf{\textcolor{highergreen}{+0.76}} && \textbf{\textcolor{highergreen}{+0.26}} & \textbf{\textcolor{highergreen}{+0.18}} & \textbf{\textcolor{highergreen}{+0.88}} & \textbf{\textcolor{highergreen}{+0.13}}\\ 
\bottomrule[1.5pt]
\end{tabular}}
\end{table}

%% file: chapters/tables/table_agedb.tex
\begin{table}[t]
% \ra{1.3}
\centering
\caption{\textbf{Results on \textit{AgeDB-DIR}.} Baseline numbers are quoted from \cite{yangetal2021}. The {best} results for each method (Vanilla, Focal-R, RRT and SQINV) are in \textbf{bold}. The best results for each metric and data subset (entire column) are in \textbf{\textcolor{bestblue}{bold and red}}.}
\label{table:agedb}

\setlength{\tabcolsep}{2.5pt}
\resizebox{0.96\linewidth}{!}{\begin{tabular}{@{}lllllcllll@{}}\toprule[1.2pt]
& \multicolumn{4}{c}{\textbf{MAE $\downarrow$}} & \phantom{abc}& \multicolumn{4}{c}{\textbf{GM $\downarrow$}} \\
\cmidrule{2-5} \cmidrule{7-10} 
& All & Man. & Med. & Few &&  All & Man. & Med. & Few \\
\midrule[1.2pt]
\textsc{Vanilla}  
& 7.77 & 6.62 & 9.55   & 13.67 && 5.05 & 4.23 & 7.01   & 10.75 \\
+ \textbf{\textsc{RankSim}} 
& {7.13}  & {6.51}   & {8.17} & {10.12} &&  {4.48} &  {4.01} &  {5.27} & 6.79\\
+ \textsc{LDS} 
& 7.67 & 6.98 & 8.86  & {10.89} && 4.85 & 4.39 & {5.80}  & {7.45} \\ 
+ \textsc{LDS} +  \textbf{\textsc{RankSim}} 
& \textbf{6.99} & \textbf{6.38} &  {7.88} &  {10.23} && \textbf{4.40} & \textbf{3.97} & {5.30} & {6.93} 
\\
+ \textsc{FDS} 
& {7.55} & {6.50} & {8.97} & {13.01} && 4.75 & 4.03 & {6.42}   & {9.93} \\ 
+ \textsc{FDS} +  \textbf{\textsc{RankSim}}    
& 7.33 &  {6.49} & 8.53 & 11.98 && 4.82 & 4.19 & 6.16 & 8.99\\
 + \textsc{LDS}  + \textsc{FDS}  
 & {7.55} & {7.01} & {8.24}   & {10.79} && {4.72} & 4.36 & {5.45}   & 6.79 \\
 + \textsc{LDS}  + \textsc{FDS}  +  \textbf{\textsc{RankSim}}  
& 7.03 & {6.54} & \textbf{7.68}  & \textbf{9.92} && 4.45 & {4.07} &  \textbf{5.23}  &  \textbf{6.35} \\
\midrule
\textsc{Focal-R}     
& 7.64 & 6.68 & 9.22 & 13.00 && 4.90 & 4.26 & 6.39   & 9.52 \\ 
 + \textbf{\textsc{RankSim}}   
 &  {7.15} & {6.45} &  {7.97}  & {11.50} &&  {4.53} & {4.10} &  {5.10}  & {8.50}  \\ 
 + {\textsc{LDS}}      
& 7.56 & {6.67} & 8.82   & 12.40 && 4.82 & 4.27 & 5.87 & 8.83 \\
 + \textsc{LDS} + \textbf{\textsc{RankSim}}   
 & 7.25 &  {6.40} & 8.71 &  {11.24} && {4.58} &  {4.02} & 5.99 & \textbf{7.52} \\
 + {\textsc{FDS}}      
& 7.65 & 6.89 & 8.70 & {11.92} && 4.83 & 4.32 & 5.89   & {8.04} \\
 + {\textsc{FDS}}  + \textbf{\textsc{RankSim}}   
 & 7.25 & 6.72 & \textbf{7.86} & \textbf{10.58} && 4.54 & 4.22 & \textbf{\textcolor{bestblue}{4.84}} &  {7.57} \\
 + {\textsc{LDS}} + {\textsc{FDS}} 
& {7.47} & 6.69 & {8.30}  & 12.55 && {4.71} & {4.25} & {5.36}  & 8.59 \\
 +  {\textsc{LDS}}  +  {\textsc{FDS}}  +\textbf{\textsc{RankSim}}   
 &\textbf{7.09} & \textbf{\textcolor{bestblue}{6.17}} & 8.71 & 11.68 && \textbf{4.46} & \textbf{\textcolor{bestblue}{3.85}} & 5.76 & 8.78 \\
\midrule
\textsc{RRT}    
& 7.74 & 6.98 & 8.79 & 11.99 && 5.00 & 4.50 & 5.88   & 8.63 \\ 
 + \textbf{\textsc{RankSim}} 
&  {7.11} &  {6.53} & {8.00}   &  {10.04} && {4.52} & {4.19} & \textbf{5.05}  & \textbf{\textcolor{black}{6.77}} \\ 
 + {\textsc{LDS}}  
& {7.72} & 7.00 & {8.75}   & {11.62} && {4.98} & 4.54 & {5.71}   & {8.27}  \\ 
 + \textsc{LDS} + \textbf{\textsc{RankSim}}    
 & \textbf{6.94} & \textbf{6.43} & \textbf{\textcolor{bestblue}{7.54}} & {10.10} && \textbf{4.37} & \textbf{3.97} &  {5.11} & {7.05} \\ 
 + {\textsc{FDS}}   
& {7.70} & {6.95} & {8.76}   & {11.86} && {4.82} & {4.32} & {5.83}   & {8.08} \\
 + \textsc{FDS} + \textbf{\textsc{RankSim}}  
  &  {7.11} & 6.55 &  {7.99} & \textbf{10.02} &&  {4.49} &  {4.13} & 5.13 &  {6.85} \\
 + {\textsc{LDS}} + {\textsc{FDS}}  
& {7.66} & 6.99 & {8.60}   & {11.32} && {4.80} & 4.42 & {5.53}   & {6.99}  \\
 + \textsc{LDS} + \textsc{FDS} + \textbf{\textsc{RankSim}}  
 & 7.13 & 6.54 & 8.07 & 10.12 && 4.55 & 4.18 & 5.20 & 6.87 \\  
 \midrule
\textsc{SQINV}      
& 7.81 & 7.16 & 8.80   & 11.20 && 4.99 & 4.57 & 5.73   & 7.77 \\
 + \textbf{\textsc{RankSim}}    
&\textbf{\textcolor{bestblue}{6.91}} & \textbf{\textcolor{black}{6.34}} & \textbf{\textcolor{black}{7.79}}   &  {9.89} && \textbf{\textcolor{bestblue}{4.28}} & \textbf{\textcolor{black}{3.92}} & \textbf{\textcolor{black}{4.88}} & 6.89 \\ 
+ {\textsc{LDS}} 
& 7.67 & {6.98} & 8.86   & 10.89 && 4.85 & {4.39} &  5.80 & {7.45}  \\ 
+ {\textsc{LDS}} + \textbf{\textsc{RankSim}}     
&  {6.99} &  {6.38} & {7.88} & 10.23 &&  {4.40} &  {3.97} & 5.30 & {6.90} \\
 + {\textsc{FDS}} 
& 7.69 & 7.10 & 8.86   & {9.98} && 4.83 & {4.41} &  5.97 & \textbf{\textcolor{bestblue}{6.29}} \\
 + {\textsc{FDS}} + \textbf{\textsc{RankSim}} 
 & 7.02 & {6.49} & {7.84} & \textbf{\textcolor{bestblue}{9.68}}  && 4.53 & {4.13} &  5.37  & {6.89} \\
 + {\textsc{LDS}} + {\textsc{FDS}} 
& {7.55} &  7.01  & {8.24}   & 10.79 && {4.72} & {4.36} & {5.45}  & {6.79} \\
 + {\textsc{LDS}} + {\textsc{FDS}} + \textbf{\textsc{RankSim}} 
 & 7.03 & {6.54} & \textbf{7.68}  & 9.92 && 4.45 & {4.07} &  {5.23}  &  {6.35} \\
\midrule
\textbf{\textsc{Ours}} \textsc{vs.} \textbf{\textsc{Vanilla}} 
& \textbf{\textcolor{highergreen}{+0.86}} & \textbf{\textcolor{highergreen}{+0.45}} & \textbf{\textcolor{highergreen}{+2.01}} & \textbf{\textcolor{highergreen}{+3.99}} && \textbf{\textcolor{highergreen}{+0.77}} & \textbf{\textcolor{highergreen}{+0.38}} & \textbf{\textcolor{highergreen}{+2.17}} & \textbf{\textcolor{highergreen}{+4.40}}  \\ 
\textbf{\textsc{Ours}} \textsc{vs.} \textbf{Yang et al. } 
& \textbf{\textcolor{highergreen}{+0.56}} & \textbf{\textcolor{highergreen}{+0.33}} & \textbf{\textcolor{highergreen}{+0.70}} & \textbf{\textcolor{highergreen}{+0.30}} && \textbf{\textcolor{highergreen}{+0.43}} & \textbf{\textcolor{highergreen}{+0.18}} & \textbf{\textcolor{highergreen}{+0.52}} & \textbf{\textcolor{worsered}{-0.06}}  \\
\bottomrule[1.5pt]
\end{tabular}}
\end{table}

%% file: chapters/tables/table_sts_short.tex
\begin{table}[t]
% \ra{1.3}
\centering
\caption{\textbf{Results on \textit{STS-B-DIR}}. %(batch size: 16, learning rate: 0.00025). 
Baseline numbers are quoted from \cite{yangetal2021}.
The {best} results for each method (Vanilla, Focal-R, RRT and INV) are in \textbf{bold}. The best results for each metric and data subset (entire column) are in \textbf{\textcolor{bestblue}{bold and red}}.}
\label{table:sts-b1}

\setlength{\tabcolsep}{2.5pt}
\resizebox{0.96\linewidth}{!}{\begin{tabular}{@{}lllllcllll@{}}\toprule[1.2pt]
& \multicolumn{4}{c}{\textbf{MSE~$\downarrow$}} & \phantom{abc}& \multicolumn{4}{c}{\textbf{Pearson cor.~(\%)~$\uparrow$}} \\
\cmidrule{2-5} \cmidrule{7-10} 
& All & Man. & Med. & Few &&  All & Man. & Med. & Few \\
\midrule[1.2pt]
\textsc{Vanilla} 
& 0.974 & 0.851 & 1.520  & 0.984 && 74.2 & 72.0 & 62.7  & 75.2  \\
 + \textbf{\textsc{RankSim}}     
& \textcolor{black}{\textbf{0.873}} & 0.908 & \textcolor{bestblue}{\textbf{0.767}}  & 0.705 &&  \textcolor{black}{\textbf{76.8}} & 71.0 & \textcolor{bestblue}{\textbf{72.9}}  & {85.2} \\
 + \textsc{LDS} 
& 0.914 &  {0.819} & 1.319 & 0.955   && 75.6 &  {73.4} & 63.8 & 76.2  \\
 + \textsc{LDS} +  \textbf{\textsc{RankSim}}
& 0.889 & 0.911 &  {0.849} &  {0.690} && 76.2 & 70.7 &  {70.0} &  {85.6} \\
 + \textsc{FDS} 
& 0.916 & 0.875 & {1.027} & 1.086  && 75.5 & 73.0 & {67.0} & 72.8 \\
 + \textsc{{FDS}}  + \textbf{\textsc{RankSim}}  
&  {0.884} & 0.924 & \textcolor{bestblue}{\textbf{0.767}} & \textcolor{black}{\textbf{0.685}} &&   {76.5} & 70.4 & 72.5 & \textcolor{black}{\textbf{85.7}}   \\ 
 + \textsc{LDS} + \textsc{FDS} 
&  {0.907} & \textcolor{bestblue}{\textbf{0.802}} & 1.363  & {0.942}    && {76.0}     & \textcolor{bestblue}{\textbf{74.0}}     & 65.2      & {76.6}  \\
 + \textsc{{LDS}} + \textsc{{FDS}} +  \textbf{\textsc{RankSim}}  
& 0.903 & 0.908 & 0.911 & 0.804   && 75.8 & 70.6 & 69.0 & 82.7  \\ 
\midrule
\textsc{Focal-R}     
& 0.951 &  {0.843} & 1.425  & 0.957 && 74.6 & 72.3 & 61.8  & 76.4 \\
 +  \textbf{\textsc{RankSim}} 
&  {0.887} & 0.889 & {0.918}  & {0.745}    &&  {76.2}     & 70.8     & {70.4}      & {84.6}  \\
 + \textsc{LDS} 
& 0.930     & \textbf{0.807}     & 1.449      & 0.993   && {75.7}     & \textbf{73.9}     & 62.4      & 75.4 \\
+ \textsc{LDS}  + \textbf{\textsc{RankSim}}
& \textbf{0.872} & 0.887 & {0.847} & \textcolor{black}{\textbf{0.718}} && \textbf{76.7} & 71.2 & 70.3 & \textbf{85.1}  \\
+ \textsc{FDS} 
& {0.920} & 0.855 & {1.169}  & 1.008  && {75.1} &  {72.6}  & {66.4} & 74.7 \\
+ \textsc{FDS}  + \textbf{\textsc{RankSim}} 
& 0.913 & 0.952 &  {0.793} &  {0.723}  && 75.6 & 69.6 & \textcolor{black}{\textbf{71.5}} &  {84.8}   \\
+ \textsc{LDS} + \textsc{FDS} 
& 0.940 & 0.849 & 1.358  & {0.916} && 74.9 & 72.2 & 66.3 & {77.3} \\
%%%%%%%%%
+ \textsc{LDS} + \textsc{FDS} + \textbf{\textsc{RankSim}} 
& 0.911 & 0.943 & \textcolor{black}{\textbf{0.779}} & 0.866 && 75.7 & 69.9 &  {71.4} & 81.2 \\
%%%%%%%%%%%%%%%%%%%%%%%%%%%
\midrule
\textsc{RRT} 
& 0.964 & 0.842 & 1.503  & 0.978 && 74.5 & 72.4 & 62.3  & 75.4 \\
 + \textbf{\textsc{RankSim}} 
& \textcolor{bestblue}{\textbf{0.865}} & {0.876} &  {0.867} & \textcolor{bestblue}{\textbf{0.670}}   &&  \textcolor{bestblue}{\textbf{77.1}} & 72.2 &  {68.3} & \textcolor{bestblue}{\textbf{86.1}}   \\
+ \textsc{LDS} 
& {0.916} &  {0.817} & {1.344}  & 0.945 && {75.7} &  {73.5} & {64.1}  & {76.6} \\
 + \textsc{LDS} + \textbf{\textsc{RankSim}} 
& 0.874 & 0.893 & \textbf{0.833} & 0.722   &&   {77.0} & 72.3 &  {68.3} & 84.8 \\
 + \textsc{FDS} 
& 0.929 & 0.857 & {1.209}  & 1.025   && 74.9     & 72.1     & {67.2}      & 74.0 \\
+ \textsc{FDS} + \textbf{\textsc{RankSim}} 
&  {0.871} & 0.874 & 0.898  & 0.734   &&  76.8 & 72.0 & \textbf{68.7} & 84.5   \\
+ \textsc{LDS} + \textsc{FDS} 
& {0.903} & \textbf{0.806} & 1.323  & {0.936}   &&  {76.0}     &\textbf{73.8}     & 65.2 & {76.7} \\
 + \textsc{LDS} + \textsc{FDS} + \textbf{\textsc{RankSim}} 
& 0.882 & 0.892 & 0.887 &  {0.702}&& 76.6 & 71.7 & 68.0 &  {85.5} \\
 \midrule
\textsc{INV}      
& 1.005 & 0.894 & 1.482  & 1.046  && 72.8 & 70.3 & 62.5  & 73.2 \\
+ \textbf{\textsc{RankSim}}  
& 1.091 & 1.056 & 1.240 & 1.118 && 69.9 & 65.2 & 60.1 & 76.0 \\
+ \textsc{LDS} 
& 0.914 &  {0.819} & {1.319}  & {0.955}  && {75.6} &  {73.4} & {63.8}  & {76.2} \\
 + \textsc{LDS}  + \textbf{\textsc{RankSim}}
& \textbf{0.889} & 0.911 & \textbf{0.849} & \textbf{0.690}  && \textbf{76.2} & 70.7 & \textbf{70.0} & \textbf{85.6}  \\
 + \textsc{FDS} 
& 0.927 & 0.851 & {1.225}  & 1.012  && 75.0     & 72.4     & {66.6} & 74.2 \\
+ \textsc{FDS}  + \textbf{\textsc{RankSim}} 
& 1.083 & 1.035 & 1.301 & 1.063 && 70.0 & 64.8 & 68.9 & 76.7  \\
 + \textsc{LDS} + \textsc{FDS} 
& {0.907} & \textcolor{bestblue}{\textbf{0.802}} & 1.363  & {0.942}  &&  {76.0} & \textcolor{bestblue}{\textbf{74.0}} & 65.2 & {76.6} \\ 
 + \textsc{LDS} + \textsc{FDS} + \textbf{\textsc{RankSim}} 
&  {0.903} & 0.908 &  {0.911} &  {0.804} &&  {75.8} & 70.6 &  {69.0} &  {82.7} \\
\midrule
\textbf{\textsc{Ours}} \textsc{vs.} \textbf{\textsc{Vanilla}} & \textbf{\textcolor{highergreen}{+0.109}} 
& \textbf{\textcolor{worsered}{-0.023}} 
& \textbf{\textcolor{highergreen}{+0.753}} 
& \textbf{\textcolor{highergreen}{+0.314}} 
&& \textbf{\textcolor{highergreen}{+2.9}} 
& \textbf{\textcolor{highergreen}{+0.3}} 
& \textbf{\textcolor{highergreen}{+10.2}} 
& \textbf{\textcolor{highergreen}{+10.9}} \\
\textbf{\textsc{Ours}} \textsc{vs.} \textbf{Yang et al. } 
& \textbf{\textcolor{highergreen}{+0.038}} 
& \textbf{\textcolor{worsered}{-0.072}} 
& \textbf{\textcolor{highergreen}{+0.260}} 
& \textbf{\textcolor{highergreen}{+0.246}} 
&& \textbf{\textcolor{highergreen}{+1.1}} 
& \textbf{\textcolor{worsered}{-1.7}} 
& \textbf{\textcolor{highergreen}{+5.7}} 
& \textbf{\textcolor{highergreen}{+8.8}} \\
\bottomrule[1.5pt]
\end{tabular}}
\vspace{-11pt}
\end{table}

%% file: chapters/ablation.tex
\subsection{Ablations and Analysis}

\input{chapters/tables/ranking_similarity}
\input{chapters/tables/feature_similarity}

\paragraph{Different choices for $\ell \,$ in Eq.~\ref{eq:RankSim}.} RankSim uses the function $\ell$ in Eq.~\ref{eq:RankSim} to penalize differences between the ranking of neighbors in label space and the ranking of neighbors in feature space. We adopted mean squared error in all of the benchmark experiments. In this ablation study, to provide a more complete picture of different ranking similarity losses, we evaluate several other options for $\ell$. Let $\mathbf{rk}^{\mathbf{a}}$ and $\mathbf{rk}^{\mathbf{b}}$ denote two $m$-dimensonal ranking vectors of interest. We consider the following options for $\ell$:

   \textsc{1) Cosine distance: } $1- \frac{\mathbf{rk}^{\mathbf{a}} \cdot \mathbf{rk}^{\mathbf{b}}}{\|\mathbf{rk}^{\mathbf{a}}\|_2 \times \|\mathbf{rk}^{\mathbf{b}}\|_2 }$

   \textsc{2) Huber loss: } $ \frac{1}{m}\sum^m_{i=1}\ell_i$, with $\delta$ default as 1.0,  \\ $\ell_i = 
     \begin{cases}
      0.5 \times (\mathbf{rk}^{\mathbf{a}}_i - \mathbf{rk}^{\mathbf{b}}_i)^2,& \text{if } |\mathbf{rk}^{\mathbf{a}}_i - \mathbf{rk}^{\mathbf{b}}_i| < \delta\\
      \delta \times (|\mathbf{rk}^{\mathbf{a}}_i - \mathbf{rk}^{\mathbf{b}}_i| - 0.5 \times \delta),& \text{otherwise}\\
    \end{cases}  $
    
   \textsc{3) $L_{\infty}$ distance: } $\max_i |\mathbf{rk}^{\mathbf{a}}_i - \mathbf{rk}^{\mathbf{b}}_i|$

   \textsc{4) MAE: } $ \frac{1}{m}\sum^m_{i=1}|\mathbf{rk}^{\mathbf{a}}_i - \mathbf{rk}^{\mathbf{b}}_i|$

   \textsc{5) MSE: } $\frac{1}{m}\sum^m_{i=1}(\mathbf{rk}^{\mathbf{a}}_i - \mathbf{rk}^{\mathbf{b}}_i)^2$

Table~\ref{table:different_distance} presents ablation results on \textit{AgeDB-DIR} and \textit{IMDB-WIKI-DIR} with SQINV. On \textit{AgeDB-DIR}, implementing RankSim with \textit{any} of the above options for $\ell$ improves the baseline SQINV. MSE achieves the best performance on the dataset overall, as well as on the many-shot subset. Cosine distance achieves the best performance on the medium-shot and few-shot subsets. On \textit{IMDB-WIKI-DIR}, cosine distance and MSE achieve comparable best performance overall and in the few-shot subset. Either option consistently improves the baseline SQINV in all settings.

\paragraph{Different choices for feature similarity function $\sigma^z$.} The similarity function $\sigma^z$ in Eq.~\ref{eq:feat_sim_matrix} quantifies the similarity of two data samples in feature space. We adopted cosine similarity in all of the benchmark experiments. Here, we evaluate RankSim with different choices for the feature-space similarity function. Let $\mathbf{z}_1$ and $\mathbf{z}_2$ denote two $d$-dimensional feature vectors. We consider the following options for $\sigma^z$:

   \textsc{1) Negative $L_{\infty}$: } $ -\max_i |\mathbf{z}_{1,i} - \mathbf{z}_{2,i}|$

   \textsc{2) Negative MAE: }
      $-{\frac{1}{d}\sum^d_{i=1} |\mathbf{z}_{1,i} - \mathbf{z}_{2,i}|}$
      
   \textsc{3) Negative MSE: }
      $ -\frac{1}{d}\sum^d_{i=1} (\mathbf{z}_{1,i} - \mathbf{z}_{2,i})^2 $
 
    \textsc{4) Correlation similarity: } $ \frac{(\mathbf{z}_1 - \overline{\mathbf{z}}_1 ) \cdot (\mathbf{z}_2 - \overline{\mathbf{z}}_2 )}{\|\mathbf{z}_1 - \overline{\mathbf{z}}_1 \|_2 \times \|\mathbf{z}_2 - \overline{\mathbf{z}}_2\|_2 }$

   \textsc{5) Cosine similarity: } $ \frac{\mathbf{z}_1 \cdot \mathbf{z}_2}{\|\mathbf{z}_1\|_2 \times \|\mathbf{z}_2\|_2 }$

Table~\ref{table:different_sigma_z} presents ablation results on \textit{AgeDB-DIR} and \textit{IMDB-WIKI-DIR} with SQINV. On \textit{AgeDB-DIR}, implementing RankSim with \textit{any} of the above options for $\sigma^z$ improves the baseline SQINV. Cosine simlarity achieves the best performance on the dataset overall, as well as on the many-shot subset. Negative MSE achieves the best performance on the few-shot subset. On \textit{IMDB-WIKI-DIR}, cosine similarity achieves the best performance overall, in the many-shot subset, and in the few-shot subset. Negative MAE, negative MSE, and cosine similarity consistently improve the baseline SQINV.

\input{chapters/tables/ablation_ties}

\paragraph{Ablation on batch sampling.} As described in Section~\ref{sec:method}, during training, we sample $\mathcal{M}$ from the current batch such that each label occurs at most once. The intent is to reduce ties and boost the relative representation of infrequent labels. In this ablation, we consider removing the sampling and using the entire batch as $\mathcal{M}$. Table~\ref{table:abltion_ties} shows the ablation results. Applying sampling leads to better performance in the few-shot and medium-shot regions, and comparable performance overall.

\begin{figure}[tp]
\centering
\includegraphics[width=0.95\linewidth]{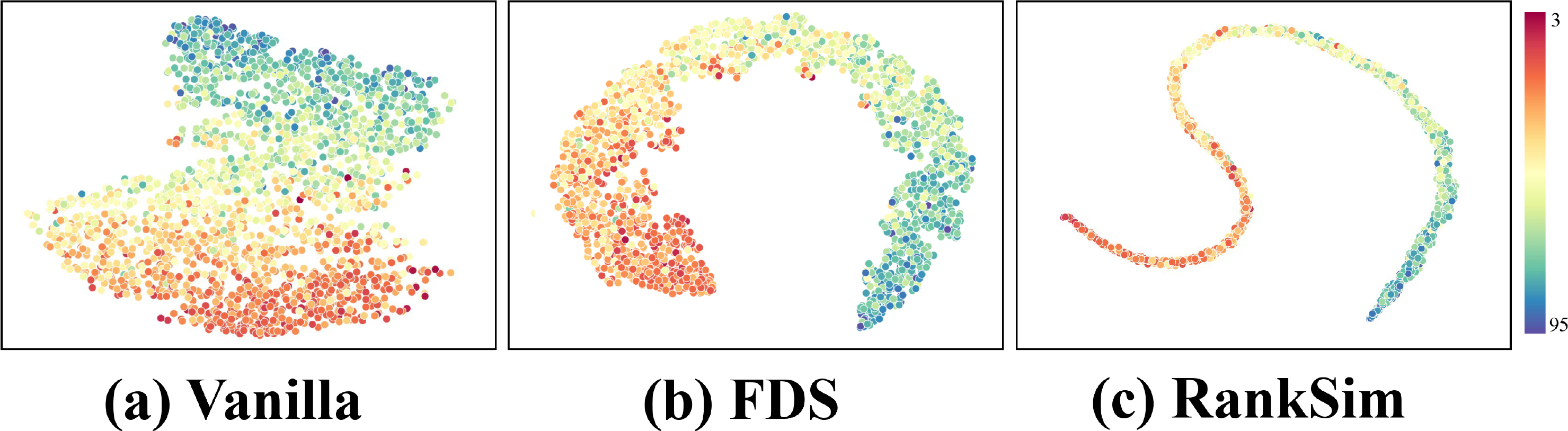}
\vspace*{-1mm}
\caption{Qualitative visualization of ResNet-50 features on \textit{AgeDB-DIR} balanced test set: \textbf{(a) Vanilla}, %\textbf{(b) LDS+FDS}, and \textbf{(c) RankSim}
\textbf{(b) Vanilla + FDS}, 
and \textbf{(c) Vanilla + RankSim}. We use t-SNE by treating the continuous labels (age, from 3 to 95) as categorical (92 ``classes'', with no samples of age 4). The label space is denoted with continuous spectral colors.} 
\label{fig:t-sne}

\end{figure}

\paragraph{Qualitative visualization of feature space.} We extract the features from ResNet-50 (Vanilla, Vanilla with FDS, and Vanilla with RankSim) on the \textit{AgeDB-DIR} test set and visualize using t-SNE technique in Fig.~\ref{fig:t-sne}. To visually differentiate labels and show the continuity in feature space, the labels are denoted in continuous spectral colors (from red to blue). The visualization shows the continuity in features learned with FDS and RankSim. 

\begin{figure}[t]
    \centering
    \includegraphics[width=1.0\linewidth]{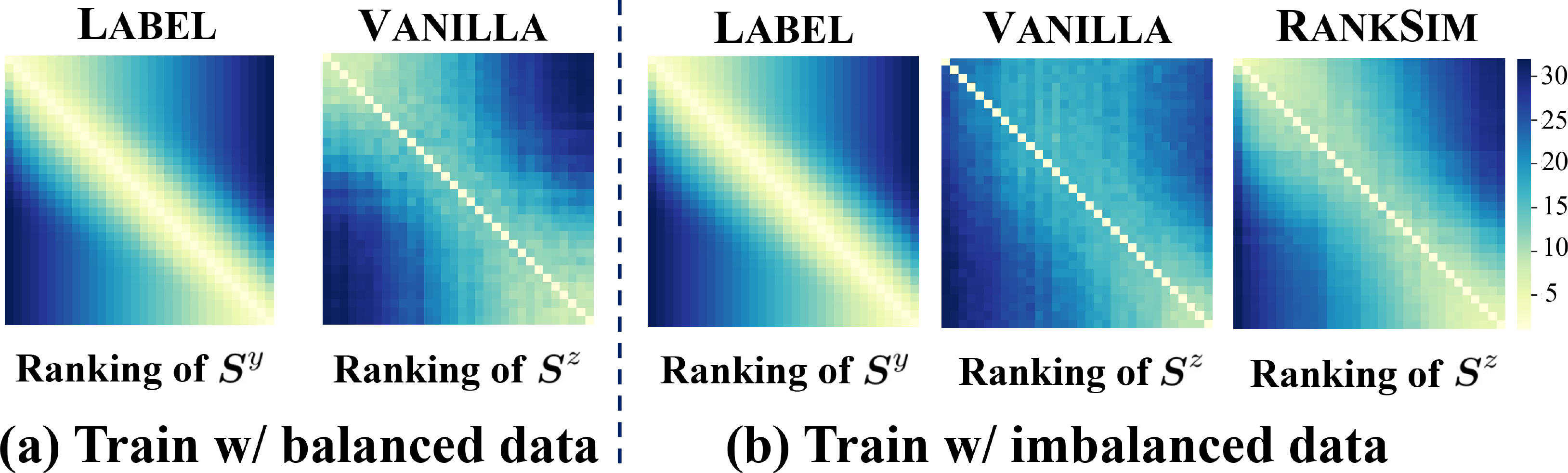}
    \vspace{-15pt}
    \caption{\textbf{Average feature-space ranking matrices $S^z$,} trained on (a) balanced subset of \textit{AgeDB-DIR} (left two), and (b) {imbalanced} full \textit{AgeDB-DIR} (right three). The corresponding label-space ranking matrix $S^y$ is provided. For clarity of visualization, each batch is pre-sorted by label.}
    \label{fig:matrix_imbalance}
\end{figure}

\paragraph{Qualitative visualization of ranking matrices with balanced and imbalanced data.} 
In Fig.~\ref{fig:matrix_imbalance}, we visualize the average (batch-wise) label-space and feature-space ranking matrices under three training settings: vanilla network on balanced data, vanilla network on imbalanced data, and vanilla with RankSim on imbalanced data. We use \textit{AgeDB-DIR} for this visualization; for the setting with balanced data, we extract a balanced subset of \textit{AgeDB-DIR} consisting of the many-shot ages only (ages in the range of 23 to 63). 
For clarity of visualization, samples in each batch are pre-sorted by label (i.e., age). Each batch consists of 32 samples. The visualized matrices are obtained by applying the ranking operation $\mathbf{rk}$ on each row of $\boldsymbol{S}^y$ and $\boldsymbol{S}^z$ as in Eq.~\ref{eq:RankSim}, and averaging the result over all test batches. For an example calculation, please refer to the supplementary. 

Fig.~\ref{fig:matrix_imbalance}(a) shows that the label-space and feature-space rankings tend to be consistent when training with balanced data. In other words, the sorted list of neighbors in label space tend to resemble the sorted list of neighbors in feature space. This validates RankSim's inductive bias. Fig.~\ref{fig:matrix_imbalance}(b) shows that this pattern disappears when training with imbalanced data. Training with the RankSim regularizer helps to recover the pattern: we can observe that the sorted list of neighbors in feature space (Fig.~\ref{fig:matrix_imbalance}(b), right) again tends to resemble the sorted list of neighbors in label space (Fig.~\ref{fig:matrix_imbalance}(b), left).

\begin{figure}[t]
\centering
\includegraphics[width=.95\linewidth]{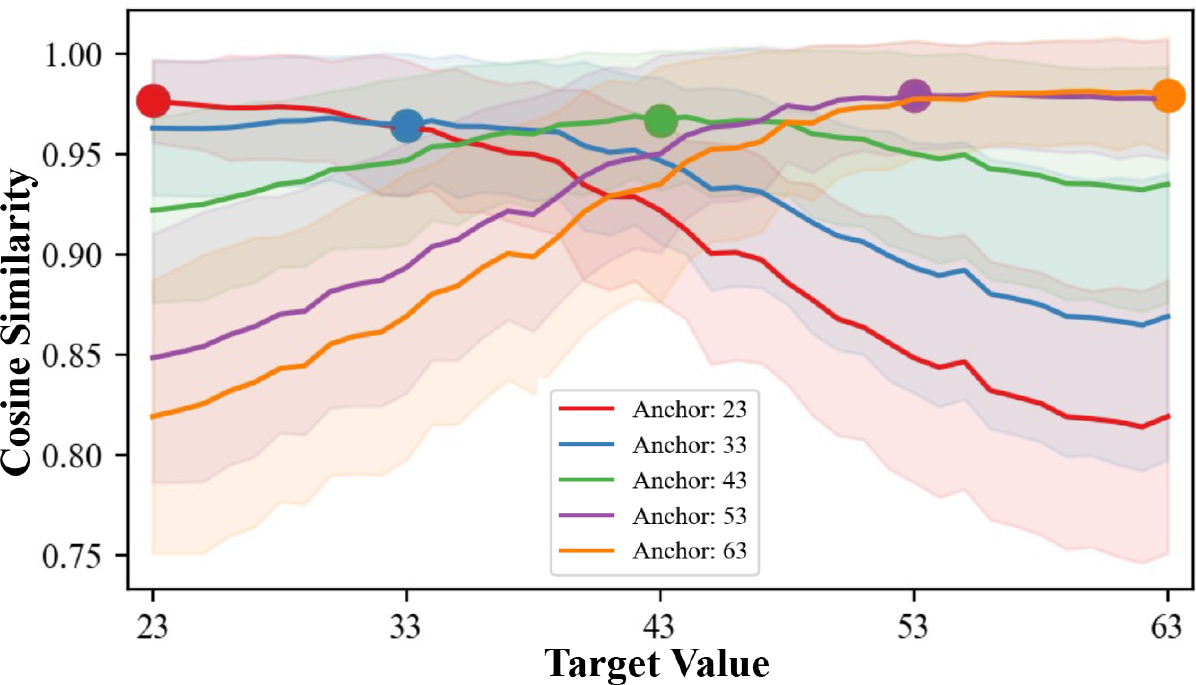}
\vspace{-1mm}
\caption{\textbf{Feature cosine similarity}, learned by Vanilla ResNet-50 from a balanced subset of \textit{AgeDB-DIR}. Each curve depicts the feature similarity between the anchor and other labels (x-axis). E.g. the point (25, 0.974) on the red curve (anchor 23) indicates the mean feature cosine similarity between age 23 and 25 is 0.974. Each solid point is the mean cosine (intra) similarity between the data points of the corresponding anchor value.}
\label{fig:balanced_curve_anchors}
\end{figure}

In Fig.~\ref{fig:balanced_curve_anchors}, we additionally plot the cosine similarity between the vanilla network features on the balanced subset.  The visualization provides an alternative view showing the consistency between label-space neighbors and feature-space neighbors when training on balanced data. For example, the curve for the youngest anchor shows a roughly monotonic drop in the average feature similarity as we move to older ages. The curve for the middle (age 43) anchor shows roughly monotonic drops in the average feature similarity in both directions as we move to younger or older ages. 
The points on each curve are the mean similarity between the data instances of the anchor itself (e.g. the red point denotes the mean cosine similarity between all data of age 23). Thus, the solid points usually have highest similarity. Fig.~\ref{fig:balanced_curve_anchors} complements Fig.~\ref{fig:matrix_imbalance}(a), showing from a feature centric view that, in the balanced setting, a vanilla network naturally learns representations such that items closer in label space are also closer in feature space.

\begin{figure}
    \centering
\includegraphics[width=0.95\linewidth]{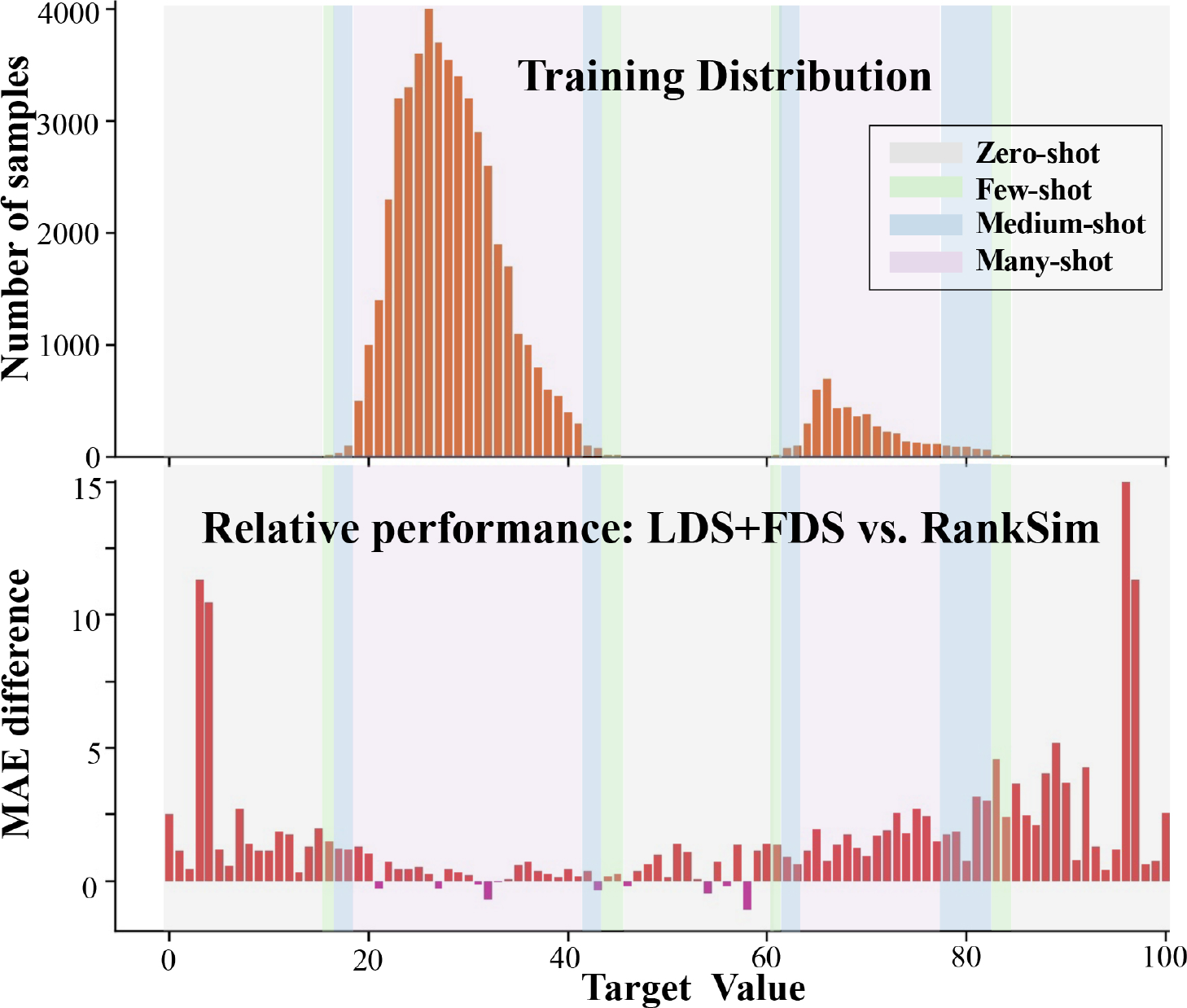}
\vspace{-1mm}
    \caption{\textbf{Training with zero-shot targets.} The top figure is a hand-crafted subset of \textit{IMDB-WIKI-DIR} that has zero-shot targets, imitating the zero-shot experiment construction in \cite{yangetal2021}. The bottom figure is the MAE difference (LDS+FDS minus RankSim) for each target value on the balanced test set. We apply SQINV re-weighting to both LDS+FDS and RankSim. In the bottom figure, y-axis $>0$ means that RankSim performs better.}
    \label{fig:mae_diff}
    \vspace{-5pt}
\end{figure}

% todo
%\paragraph{Training curve of feature distance.} We can visualize the training curve of feature distance. We can choose an anchor and show its averaged distance to other labels in both balanced training and imbalanaced training. 
\paragraph{Zero-shot targets.} 
To demonstrate the ability of RankSim to handle the zero-shot scenario, we construct a subset of \textit{IMDB-WIKI-DIR} that contains zero-shot targets. Fig.~\ref{fig:mae_diff} (top) shows the hand-crafted distribution, which is intended to imitate the zero-shot experiment construction in \citet{yangetal2021}. Fig.~\ref{fig:mae_diff} (bottom) shows that, in addition to performing comparably with LDS+FDS in or near the many-shot region, RankSim significantly outperforms LDS+FDS in the zero-shot region. The performance gap is largest for the zero-shot targets at the extremes. We surmise that the global ranking-driven representation learned by RankSim is effective at both interpolation and extrapolation, while LDS and FDS, as local smoothing-based methods, are primarily designed for interpolation.

%\paragraph{Different hyperparameters for RankSim.}
\paragraph{Training cost.} We measured the training time for {\textit{AgeDB-DIR}} on four NVIDIA GeForce GTX 1080 Ti GPUs. RankSim incurs a small overhead with respect to vanilla training and is faster to train than FDS: for one epoch, vanilla takes 12.2 seconds for a forward pass and 31.5 seconds for training; RankSim takes 16.8 seconds for a forward pass and 38.8 seconds for training; FDS takes 38.4 seconds for a forward pass and 60.5 seconds for training. 

\paragraph{Sensitivity to hyperparameters.}
RankSim introduces two hyperparameters: the balancing weight $\gamma$ and the interpolation strength $\lambda$. 
We performed sensitivity experiments on \textit{AgeDB-DIR} and \textit{IMDB-WIKI-DIR} in which we varied these two hyperparameters. The full results can be found in the supplementary. Performance is relatively robust to both $\gamma$ and $\lambda$. 
Since RankSim requires a batch-wise calculation to construct the pairwise similarity matrices, we include in the supplementary additional sensitivity experiments in which we varied the batch size.

%% file: chapters/tables/ranking_similarity.tex
\begin{table}[t]
\setlength{\tabcolsep}{2.5pt}
\caption{\textbf{Different choices of ranking similarity penalty function $\ell$} in Eq.~\ref{eq:RankSim}. Results are on \textit{AgeDB-DIR} (top) and \textit{IMDB-WIKI-DIR} (bottom) with SQINV. MSE is used in all other experiments.
}
\vspace{-4pt}
\label{table:different_distance}
\small
\vskip 0.15in
\begin{center}
\resizebox{0.48\textwidth}{!}{

\begin{tabular}{@{}lllllcllll@{}}\toprule[1.2pt]
\multicolumn{1}{c}{\multirow{2}{*}{\textbf{\textit{AgeDB-DIR}}}} & \multicolumn{4}{c}{\textbf{MAE $\downarrow$}} & \phantom{abc}& \multicolumn{4}{c}{\textbf{GM $\downarrow$}} \\

\cmidrule{2-5} \cmidrule{7-10}
    & All   & Many  & Med. & Few   && All   & Many   & Med. & Few\\ %& All   & Many  & Med. & Few   \\ 
 \midrule
 \textsc{Cosine distance} & 6.99 & 6.55 & \textbf{\textcolor{black}{7.61}} & \textbf{\textcolor{black}{9.47}} && 4.40 & 4.14 & \textbf{4.84} & \textbf{5.95} \\%& 85.00 & 74.08 & \textbf{99.23} & \textbf{148.77}\\
\textsc{Huber} %loss 
& 7.05 & 6.43 & 7.99 & 10.25 && 4.53 & 4.07 & 5.68 & 6.69 \\
\textsc{$L_{\infty}$} %loss 
& 7.07 & 6.53 & 7.81 & 10.08 && 4.60 & 4.27 & 5.07 & 7.06 \\%& 85.21 & 72.24 & 101.31 & 163.17\\
\textsc{MAE} %loss 
& 7.04 & 6.42 & 7.94 & 10.42 && 4.37 & 3.95 & 5.18 & 7.27 \\%& 84.63 & 71.62 & 101.29 & 161.43 \\
\textsc{MSE} %loss 
& \textbf{\textcolor{black}{6.91}} & \textbf{\textcolor{black}{6.34}} & {7.79}   & {9.89} && \textbf{\textcolor{black}{4.28}} & \textbf{\textcolor{black}{3.92}} & 4.88 & {6.89}\\ %& \textbf{82.10} & \textbf{68.60} & {102.61} & {152.84}\\
\midrule[1.2pt]

\multicolumn{1}{c}{\multirow{2}{*}{\textbf{\textit{IMDB-WIKI-DIR}}}} & \multicolumn{4}{c}{\textbf{MAE $\downarrow$}} & \phantom{abc}& \multicolumn{4}{c}{\textbf{GM $\downarrow$}} \\
\cmidrule{2-5} \cmidrule{7-10}
    & All   & Many  & Med. & Few   && All   & Many   & Med. & Few\\ %& All   & Many  & Med. & Few   \\ 
 \midrule
\textsc{Cosine distance}
& \textbf{7.38} & \textbf{6.79} & 12.25 & \textbf{21.87} && \textbf{4.07} & \textbf{3.83} & 7.14 & 13.62 \\% & \textbf{123.42} & \textbf{100.57} & 298.58 & \textbf{785.30} \\
\textsc{Huber} %loss 
& 7.46 & 6.87 & 12.22 & 22.87 && 4.17 & 3.94 & 6.87 & 14.07 \\%& 124.43 & 100.64 & 301.27 & 857.23 \\ 
$L_{\infty}$ %loss 
& 7.57 & 7.01 & \textbf{11.99} & 22.28 && 4.29 & 4.08 & \textbf{6.54} & 13.59\\% & 126.16 & 103.82 & 292.89 & 810.00\\
MAE %loss 
 & 7.48 & 6.86 & 12.50 & 23.13 && 4.14 & 3.89 & 7.07 & 14.79 \\%& 126.70 & 102.29 & 310.28 & 862.61\\
MSE %loss 
& 7.42 & {6.84} & 12.12 & 22.13 && {4.10} & {3.87} & 6.74 &  \textbf{12.78} \\ %& \textbf{123.76} & {101.02} & {296.73} & \textbf{793.55}\\
\bottomrule[1.2pt]
\end{tabular}
}
\end{center}
%\vspace{-0.5cm}
\end{table}

%% file: chapters/tables/feature_similarity.tex
\begin{table}[t]
\setlength{\tabcolsep}{2.5pt}
\caption{\textbf{Different choices of feature similarity function $\sigma^z$.} Results are on \textit{AgeDB-DIR} (top) and \textit{IMDB-WIKI-DIR} (bottom) with SQINV. Cosine similarity is used in all other experiments.}
\vspace{-4pt}
\label{table:different_sigma_z}
\small
\vskip 0.15in
\begin{center}
\resizebox{0.48\textwidth}{!}{
% \begin{tabular}{lcccccccc} %@{}lllllcrrrr@{}
% \toprule[1.5pt]

\begin{tabular}{@{}lllllcllll@{}}\toprule[1.2pt]
\multicolumn{1}{c}{\multirow{2}{*}{\textbf{\textit{AgeDB-DIR}}}} & \multicolumn{4}{c}{\textbf{MAE $\downarrow$}} & \phantom{abc}& \multicolumn{4}{c}{\textbf{GM $\downarrow$}} \\

% \multicolumn{1}{c}{\multirow{2}{*}{\textbf{\textit{AgeDB-DIR}}}}    & \multicolumn{4}{c}{\textbf{MAE~$\downarrow$}}       & \multicolumn{4}{c}{\textbf{GM~$\downarrow$}}  \\ %& \multicolumn{4}{c}{\textbf{MSE~$\downarrow$}}  \\
\cmidrule{2-5} \cmidrule{7-10}
    & All   & Many  & Med. & Few   && All   & Many   & Med. & Few\\ %& All   & Many  & Med. & Few   \\ 
 \midrule
\textsc{Negative $L_{\infty}$}
& {7.14} & {6.56} & {7.92} & {10.37} && {4.51} & {4.15} & 5.11 & 7.05\\ %& 88.32 & 73.96 & 105.87 & 175.43\\
\textsc{Negative MAE}  
& {7.25} & {6.54} & 8.40 & {10.77} && {4.67} & {4.18} & 5.64 & 7.89\\ %& 88.17 & 72.08 & 112.64 & 172.48 \\
\textsc{Negative MSE}  
& 7.06 & 6.61 & \textbf{7.80} & \textbf{9.20} && {4.43} & 4.17 & 4.95 & \textbf{5.83}\\ %& {84.72} & {73.71} & \textbf{101.31} & \textbf{142.84} \\
\textsc{Correlation similarity} 
& {7.32} & {6.77} & 8.24 & {10.01} && {4.75} & {4.43} & 5.37 & 6.44\\ %& 91.12 & 75.85 & 114.88 & 169.69 \\ 
\textsc{Cosine similarity} 
& \textbf{6.91} & \textbf{6.34} & {7.79}   & {9.89} && \textbf{4.28} & \textbf{3.92} & \textbf{4.88} & {6.89}\\ %& \textbf{82.10} & \textbf{68.60} & {102.61} & {152.84}\\
\midrule[1.2pt]
\multicolumn{1}{c}{\multirow{2}{*}{\textbf{\textit{IMDB-WIKI-DIR}}}} & \multicolumn{4}{c}{\textbf{MAE $\downarrow$}} & \phantom{abc}& \multicolumn{4}{c}{\textbf{GM $\downarrow$}} \\
\cmidrule{2-5} \cmidrule{7-10}
    & All   & Many  & Med. & Few   && All   & Many   & Med. & Few\\ %& All   & Many  & Med. & Few   \\ 
 \midrule
\textsc{Negative $L_{\infty}$} 
& 7.44 & 6.85 & 12.10 & 23.33 && 4.15 & 3.92 & 6.61 & 15.14\\ %& 124.97 & 101.26 & 298.59 & 876.10\\
\textsc{Negative MAE}   
& 7.48 & 6.92 & \textbf{11.86} & 22.56 && 4.21 & 3.99 & \textbf{6.56} & 14.33\\ %& 124.11 & 101.79 & \textbf{288.23} & 826.65\\
\textsc{Negative MSE}    
& 7.46 & 6.87 & 12.26 & 22.20 && 4.17 & 3.93 & 6.97 & 14.55\\ %& 124.02 & \textbf{100.72} & 302.36 & 800.03\\
\textsc{Correlation similarity}
& 7.78 & 7.03 & 14.09 & 24.97 && 4.26 & 3.93 & 9.00 & 16.59\\ %& 133.40 & 105.78 & 345.29 & 932.09\\
% & 7.54 & 6.94 & 12.38 & 23.10 & 4.18 & 3.93 & 7.21 & 14.70 & 125.33 & 102.26 & 294.43 & 855.35 \\ 
\textsc{Cosine similarity} 
& \textcolor{black}{\textbf{7.42}} & \textbf{6.84} & {12.12} & \textbf{22.13} && \textbf{4.10} & \textbf{3.87} & {6.74} &  \textbf{12.78}\\ %& \textbf{123.76} & {101.02} & {296.73} & \textbf{793.55}\\
\bottomrule[1.2pt]
\end{tabular}
}
\end{center}
%\vspace{-0.5cm}
\end{table}

%% file: chapters/tables/ablation_ties.tex
\begin{table}[t]
\setlength{\tabcolsep}{2.5pt}
\caption{\textbf{Ablation study on batch sampling.} ``With sampling" refers to sampling $\mathcal{M}$ from the current batch such that each label occurs at most once. Results are on \textit{IMDB-WIKI-DIR} and \textit{AgeDB-DIR} with SQINV re-weighting, and \textit{STS-B-DIR} with Vanilla. We apply sampling in all other experiments.}
\vspace{-4pt}
\label{table:abltion_ties}
\small
\vskip 0.15in
\begin{center}
\resizebox{0.48\textwidth}{!}{

\begin{tabular}{@{}lllllcllll@{}}\toprule[1.2pt]
\multicolumn{1}{c}{\multirow{2}{*}{\textbf{\textit{AgeDB-DIR}}}} & \multicolumn{4}{c}{\textbf{MAE $\downarrow$}} & \phantom{abc}& \multicolumn{4}{c}{\textbf{GM $\downarrow$}} \\

\cmidrule{2-5} \cmidrule{7-10}
    & All   & Many  & Med. & Few   && All   & Many   & Med. & Few\\
 \midrule
\textsc{with Sampling} 
& \textbf{6.91} & \textbf{6.34} & \textbf{7.79}   & \textbf{9.89} && \textbf{4.28} & \textbf{3.92} & \textbf{4.88} & {6.89} \\
\textsc{without Sampling} 
& 7.10 & 6.52 & 7.99 & 10.17 && 4.53 & 4.17 & 5.16 & \textbf{6.85} \\ 
\midrule[1.2pt]
\multicolumn{1}{c}{\multirow{2}{*}{\textbf{\textit{IMDB-WIKI-DIR}}}} & \multicolumn{4}{c}{\textbf{MAE $\downarrow$}} & \phantom{abc}& \multicolumn{4}{c}{\textbf{GM $\downarrow$}} \\
\cmidrule{2-5} \cmidrule{7-10}
    & All   & Many  & Med. & Few   && All   & Many   & Med. & Few\\ 
 \midrule
\textsc{with Sampling} 
& \textcolor{black}{\textbf{7.42}} & \textbf{6.84} & \textbf{12.12} & \textbf{22.13} && \textbf{4.10} & \textbf{3.87} & \textbf{6.74} &  \textbf{12.78} \\
\textsc{without Sampling} 
& 7.45 & {6.85} & {12.27} & 23.05 && {4.14} & {3.89} & {7.04} &  {15.67} \\
\midrule[1.2pt]
\multicolumn{1}{c}{\multirow{2}{*}{\textbf{\textit{STS-B-DIR}}}} & \multicolumn{4}{c}{\textbf{MSE $\downarrow$}} & \phantom{abc}& \multicolumn{4}{c}{\textbf{Pearson cor.~(\%)~$\uparrow$}}  \\
\cmidrule{2-5} \cmidrule{7-10}
    & All   & Many  & Med. & Few   && All   & Many   & Med. & Few\\ 
 \midrule
\textsc{with Sampling} 
& {0.873} & 0.908 & \textbf{0.767}  & \textbf{0.705}&
& \textbf{76.8} & 71.0 & \textbf{72.9}  & \textbf{85.2} \\
\textsc{without Sampling} 
& \textbf{0.869} & \textbf{0.869} & {0.884} & 0.805& 
& 76.7 & \textbf{71.6} & 69.1 & 83.3 \\
\bottomrule[1.2pt]
\end{tabular}
}
\end{center}
%\vspace{-0.5cm}
\end{table}

%% file: chapters/conclusion.tex
\section{Conclusion}

We introduced a novel regularizer for deep imbalanced regression that captures both nearby and distant relationships in label and feature space. RankSim achieves new state-of-the-art results on three benchmarks for deep imbalanced regression, and is complementary to conventional imbalanced learning techniques, including re-weighting,  two-stage (decoupled) training, and label and feature distribution smoothing. We hope that our findings will unlock further advances in the training of regression networks with imbalanced data.

\paragraph{Limitations and future work.} We have focused on training scenarios where the data samples are independent. In some applications, such as dense pixelwise regression in computer vision, inputs are coupled (e.g. pixels in an image are not independent) and the translation of the concept of neighbors in label space and feature space may require additional work. We leave this for a future extension. In addition, we plan to further study the generalization of the RankSim regularizer to representation learning more broadly.

%% file: chapters/supp.tex
\begin{center}
\LARGE{\textbf{Supplementary Material}}
\end{center}

\section{Baseline Details}\label{supp:mode_baseline}
In this section, we provide a detailed description of all models including RankSim and all other baselines. In Table~\ref{table:imdb-wiki-dir}, Table~\ref{table:agedb} and Table~\ref{table:sts-b1}, we apply RankSim to regularize a variety of standard imbalanced learning methods. We also apply RankSim to the state-of-the-art LDS and FDS methods~\cite{yangetal2021}, and follow them to combine their method with other standard methods. As RankSim is orthogonal to conventional imbalanced learning techniques, we apply it to all baselines to provide a full picture of its effectiveness. 
% \subsection{Baselines}

\textbf{\textsc{Vanilla}.} It refers to conventionally train the backbone network without any imbalanced learning methods, i.e. ResNet-50 for \textit{IMDB-WIKI-DIR} and \textit{AgeDB-DIR}, and BiLSTM+GloVe for \textit{STS-B-DIR}. 

\textbf{\textsc{Focal-R}.} Inspired by Focal loss~\cite{lintsungyi2018} which adds a factor to standard cross entropy to focus on hard samples, \citet{yangetal2021} proposed its regression version Focal-R: $\frac{1}{n}\sum^n_{i=1}\operatorname{Sigmoid}(| \beta e_i |)^\gamma$, where $e_i$ is the error for $i$-th sample, $\beta$ and $\gamma$ are hyperparameters. The scaling factor is now a continuous function that maps the absolute error into the range of [0,1]. It shares the same idea as Focal loss to focus on hard samples. $L_1$ distance is used for $e_i$ in all experiments with Focal-R in this paper. $\beta$ and $\gamma$ are set to 0.2 and 1.0 for all experiments. 

\textbf{\textsc{RRT}.} \citet{yangetal2021} extended the standard re-training method to regression by decoupling the training of backbone feature extractor and regressor (originally classifier~\cite{Kang2020Decoupling}) into two stages, and termed it as Regressor Re-training (RRT). In the first stage, the network is trained normally, and in the second stage only the regressor (the last FC layer) is re-trained with inverse re-weighting. As RankSim is a regularizer for the feature learning, we apply it in the first stage. 

\textbf{\textsc{SQInv} and \textsc{Inv}.} The cost-sensitive re-weighting methods can be used for imbalanced regression by first dividing the target space into a finite number of bins. By calculating the number of samples in each bin, standard imbalanced classification method can be adopted. To fairly compare with state-of-the-art experiments, we also use square-root inverse-frequency weighting (SQINV) for \textit{IMDB-WIKI-DIR} and \textit{AgeDB-DIR}, and inverse-frequency weighting (INV) for \textit{STS-B-DIR}.

\textbf{\textsc{LDS}.} Label Distribution Smoothing~\cite{yangetal2021} convolves a symmetric kernel with the empirical label distribution to account for the continuity of labels. It has two hyperparameters---the kernel size and the standard deviation. In all experiments with LDS in this paper, we use Gaussian kernel with the validated hyperparameters in the original papers (the kernel size as 5 and the standard deviation as 2).  

\textbf{\textsc{FDS}.} Feature Distribution Smoothing~\cite{yangetal2021} performs distribution smoothing on the feature space. It transfers the feature statistics between nearby target bins. Similar to LDS, it has two hyperparameters---the kernel size and the standard deviation. In all experiments with FDS in this paper, we use Gaussian kernel with the validated hyperparameters in the original papers (the kernel size as 5 and the standard deviation as 2).

\section{Experiment Details}
In this section, we provide the full details about training details and additional experimental results. We train all models with 4 GPUs. We directly use all standard train/val/test splits created by~\citet{yangetal2021} for fair comparison. The details about training are provided in the subsections below.

\subsection{Full Evaluation Metrics} In the main manuscript, we report MAE and GM for \textit{IMDB-WIKI-DIR} and \textit{AgeDB-DIR}, and report MSE and Pearson correlation for \textit{STS-B-DIR}. We provide more metrics to evaluate RankSim and other baselines.

\textbf{MAE.} Mean absolute error (MAE) is defined as $\frac{1}{N}\sum^N_{i=1}|y_i-\Hat{y}_i|$. For $i$-th sample, $y_i$ is the ground truth label, $\Hat{y}_i$ is the prediction, and $N$ is the number of samples. Lower is better.

\textbf{MSE.} Mean squared error (MSE) is defined as $\frac{1}{N}\sum^N_{i=1}(y_i-\Hat{y}_i)^2$. For $i$-th sample, $y_i$ is the ground truth label, $\Hat{y}_i$ is the prediction, and $N$ is the number of samples. Lower is better. 

\textbf{GM.} Geometric Mean (GM) is defined as $(\prod^N_{i=1}|y_i - \Hat{y}_i|)^{\frac{1}{N}}$. For $i$-th sample, $y_i$ is the ground truth label, $\Hat{y}_i$ is the prediction, and $N$ is the number of samples. Lower is better.

\textbf{Pearson correlation.} Following the original evaluation on \textit{STS-B}, we use Pearson correlation, defined as $\frac{\operatorname{cov}(\mathbf{y}, \Hat{\mathbf{y}})}{\sigma_\mathbf{y}\sigma_\mathbf{\Hat{y}}}$, to evaluate the linear relationship between predictions and
 ground truth labels. Please refer to the GLUE benchmark~\cite{wang-etal-2018-glue} for more details. Higher is better.

\textbf{Spearman correlation.} Following the original evaluation on \textit{STS-B}, we use Spearman correlation, defined as the Pearson correlation between the rank variables, to evaluate the linear relationship between predictions and
 ground truth labels. Please refer to the GLUE benchmark~\cite{wang-etal-2018-glue} for more details. Higher is better.

\subsection{Experiments on \textit{IMDB-WIKI-DIR}}
In Table~\ref{table:imdb-wiki-dir}, we report the experimental results on \textit{IMDB-WIKI-DIR} with MAE and GM. In this section, we further report additional evaluation metrics. 

\textbf{Training details.} For all experiments, we use ResNet-50 as the backbone network. We use Adam optimizer with 0.9 momentum and 1e-4 weight decay. The learning rate is 1e-3 and the batch size is 256.
We train all methods for 90 epochs with The learning rate scheduled to drop by 10 times at epoch 60 and epoch 80. The input image size is 224 by 224. For RankSim hyperparameters, we set the balancing weight $\gamma$ as 100 and the interpolation strength $\lambda$ as 2 for all experiments on \textit{IMDB-WIKI-DIR}. In all experiments, the best epoch is selected based on the `\textit{All}' MAE on the validation set.

\input{chapters/tables/supp_imdb}
\subsection{Experiments on \textit{AgeDB-DIR}}

\paragraph{Training details}
In the main manuscript, we reported RankSim results trained with batch size of 64 and learning rate of 2.5e-4, and quoted the baseline numbers from the original paper~\cite{yangetal2021}. We found that this batch size and learning rate can also often produce better baseline results. To provide a full comparison, we include these additional baseline results in Table~\ref{table:agedb_supp} in brackets. RankSim outperforms the baselines in both hyperparameter settings.
We train all methods for 90 epochs with Adam optimizer. The momentum is 0.9 and the weight decay is 1e-4. The learning rate scheduled to drop by 10 times at epoch 60 and epoch 80. The input image size is 224 by 224. 
For RankSim hyperparameters, we set the balancing weight $\gamma$ as 100 and the interpolation strength $\lambda$ as 2 for all experiments (for both batch size of 64 and 256) on \textit{AgeDB-DIR}. In all experiments, the best epoch is selected based on the `\textit{All}' MAE on the validation set.

\input{chapters/tables/supp_agedb}

\subsection{Experiments on \textit{STS-B-DIR}}

\paragraph{Training details}
We follow the same training details as \citet{yangetal2021}---train with Adam optimizer, validate the model every 10 epochs, use MSE as the validation metric, and stop training
when the validation error does not decrease after 10 validation checks.
We use a smaller batch size of 16 and learning rate of 2.5e-4 as these produce better results (lower `\textit{All}' MSE error) on the validation set. We set the maximum number of validation checks as 300, and patience for early stopping as 30. 
For RankSim hyperparameters, we set the balancing weight $\gamma$ as 3e-4 and the interpolation strength $\lambda$ as 2 for all experiments on \textit{STS-B-DIR}.

\input{chapters/tables/supp_sts}

\section{Ablation Studies and Analysis}
\subsection{Different loss functions for ranking similarity}

We provide full details of the ablation study on the ranking similarity loss function $\ell$ in Eq.~\ref{eq:RankSim}. We consider five loss functions---mean squared error (MSE), mean absolute error (MAE), cosine distance, huber loss, and $L_\infty$ loss. Please refer to the main manuscript for the mathematical formulation of each function. In all other experiments in this paper, we keep this ranking similarity loss function as MSE. We conduct experiments on AgeDB-DIR dataset and choose ResNet-50 as backbone network. Beside the cosine similarity we used in other experiments, in this experiment we finetune RankSim with other ranking similarity loss functions. For the hyperparameters of RankSim,  we set cosine distance, :$\gamma$ as 1 and $\lambda$ as 2; Huber loss function: $\gamma$ as 1 and $\lambda$ as 2; $L_\infty$ loss: $\gamma$ as 500 and $\lambda$ as 2; MAE loss: $\gamma$ as 200 and $\lambda$ as 2.

\subsection{Different similarity functions for feature similarity}

We provide full details of the ablation study on the feature similarity function $\sigma$ in Eq.~\ref{eq:feat_sim_matrix}. We consider five similarity functions---cosine similarity, correlation similarity, MSE similarity, MAE similarity, and $L_\infty$ similarity. As this function is a similarity function, for MSE/MAE/$L_\infty$ similarity, we simply negate the loss function of MSE/MAE/$L_\infty$ to be its similarity function. Please refer to the main manuscript for the mathematical formulation of each function. In all other experiments in this paper, we keep this feature similarity function as cosine similarity. We conduct experiments on AgeDB-DIR dataset and choose ResNet-50 as backbone network. Beside the cosine similarity we used in other experiments, in this experiment we finetune RankSim with other feature similarity functions. For the hyperparameters of RankSim,  we set correlation similarity function, :$\gamma$ as 0.1 and $\lambda$ as 2; MSE similarity function: $\gamma$ as 50 and $\lambda$ as 2; MAE similarity function: $\gamma$ as 200 and $\lambda$ as 2; $L_\infty$ similarity function: $\gamma$ as 500 and $\lambda$ as 2. 

\subsection{Qualitative visualization of ranking matrices}

The visualized matrices are obtained by applying the ranking operation $\mathbf{rk}$ on each row of $\boldsymbol{S}^y$ and $\boldsymbol{S}^z$ as in Eq.~\ref{eq:RankSim}, and averaging the result over all test batches. For example, row 1 in the matrix indicates the averaged rankings with respect to the first (youngest) data sample in a batch. Suppose we have a toy test set consisting of the batch in Fig.~\ref{fig:illustration} ($y = [1, 21, 25, 70]$ and $\mathbf{rk}(\boldsymbol{S}^z_{[1,:]}) = [1, 3, 4, 2]$), plus one additional batch in which $y = [5, 10, 40, 56]$ and $\mathbf{rk}(\boldsymbol{S}^z_{[1,:]}) = [1, 2, 4, 3]$. Then row 1 of the average feature-space ranking matrix would be given by the average of $[1, 3, 4, 2]$ and $[1, 2, 4, 3]$: that is, $[1, 2.5, 4, 2.5]$.

\subsection{Sensitivity to hyperparameters}
In this subsection, we present sensitivity experiments on the two hyperparameters of RankSim (the balancing weight $\gamma$ and the interpolation strength $\lambda$) on \textit{AgeDB-DIR} in Table~\ref{table:hyperparams_agedb} and \textit{IMDB-WIKI-DIR} in Table~\ref{table:hyperparams_imdb}.

\input{chapters/tables/hyperparameter}
\newpage
\subsection{Varying batch size}

Since RankSim requires a batch-wise calculation to construct the pairwise similarity matrices, we also conducted experiments in which we varied the batch size.
\input{chapters/tables/ablation_batchsize}

%% file: chapters/tables/supp_imdb.tex
\begin{table}[h]
\caption{Complete Results on \textit{IMDB-WIKI-DIR}. Baseline numbers are quoted from \cite{yangetal2021}. The {best} results for each method (Vanilla, Focal-R, RRT, SQINV) are in \textbf{bold}. The best results for each metric and data subset (entire column) are in \textbf{\textcolor{bestblue}{bold and red}}.}% We can observe RankSim can significantly improve all methods.}
% \textcolor{red}{Same as original hyperparameters.}}
\label{table:imdb-wiki-dir-complete}
\setlength{\tabcolsep}{2.5pt}
%\small
%\vskip 0.15in
\begin{center}
\resizebox{0.9\linewidth}{!}{
\begin{tabular}{l|cccc|cccc|cccc}
\toprule[1.2pt]
\textbf{Metrics} & \multicolumn{4}{c|}{\textbf{MAE $\downarrow$}} & \multicolumn{4}{c}{\textbf{GM $\downarrow$}}  & \multicolumn{4}{c}{\textbf{MSE $\downarrow$}}\\
\midrule
\textbf{Shot} & All & Many & Med. & Few & All & Many & Med. & Few & All & Many & Med. & Few\\
\midrule
\midrule
\textsc{Vanilla} & 8.06 & 7.23 & 15.12 & 26.33 & 4.57 & 4.17 & 10.59 & 20.46  &138.06 &108.70 & 366.09 & 964.92 \\
\textsc{Vanilla} + \textsc{LDS}  & 7.83 & 7.31 & {12.43} & 22.51 & 4.42 & 4.19 & {7.00} & 13.94 & 131.65 & 109.04 & {298.98} & 829.35 \\
\textsc{Vanilla} + \textsc{FDS}  &  7.85 & {7.18} & 13.35 & 24.12 & 4.47 & 4.18 & 8.18 & 15.18 & 133.81 & 107.51 & 332.90 & 916.18 \\
\textsc{Vanilla} + \textsc{LDS} + \textsc{FDS} & {7.78} & {7.20} & {12.61} & {22.19} & {4.37} & {4.12} & {7.39} & {12.61} & {129.35} & {106.52} & 311.49 & {811.82} \\\midrule
\textsc{Vanilla} + \textbf{\textsc{RankSim}}  
& 7.72 & \textbf{6.93} & 14.48 & 25.38 & {4.27} & \textbf{3.90} & 10.02 & {17.84} & 130.60 & 102.88 & 343.01 & 934.21\\
\textsc{Vanilla} + \textsc{LDS} + \textbf{\textsc{RankSim}}  
& \textbf{7.57} & {7.00} & \textbf{12.16} & \textbf{22.44} & \textbf{4.23} & {4.00} & \textbf{6.81} & 13.23 & \textbf{127.29} & {104.07} & \textbf{299.81} & 844.00 \\
\textsc{Vanilla} + \textsc{FDS} + \textbf{\textsc{RankSim}} & {7.74} & \textbf{6.93} & {14.71} & {24.91} & {4.34} & {3.96} & {10.35} & {16.85} & {129.78} & \textbf{101.21} & 350.91 & {940.71}\\
\textsc{Vanilla} + \textsc{LDS} + \textsc{FDS} + \textbf{\textsc{RankSim}} 
& \textbf{7.69} & {7.13} & {12.30} & \textcolor{bestblue}{\textbf{21.43}} & {4.34} & 4.13 & {6.72} &  \textcolor{bestblue}{\textbf{12.48}} & {129.09} & {106.21} & {303.82} & \textbf{798.05} \\
\midrule\midrule
\textsc{Focal-R}
& 7.97 & 7.12 & 15.14 & 26.96 & 4.49 & 4.10 & 10.37 & 21.20 & 136.98 & 106.87 & 368.60 & 1002.90\\
\textsc{Focal-R} + LDS  
& 7.90 & 7.10 & {14.72} & 25.84 & 4.47 & {4.09} & {10.11} & 19.14 & {132.81} & 105.62 & {354.37} & {949.03} \\
\textsc{Focal-R} + FDS  
& 7.96 & 7.14 & {14.71} & 26.06 & 4.51 & 4.12 & {10.16} & 19.56 & 133.74 & 105.35 & 351.00 & 958.91 \\
\textsc{Focal-R} + LDS + FDS 
& {7.88} & {7.10} & {14.08} & {25.75} & {4.47} & 4.11 & {9.32} & {18.67}& {132.58} & {105.33} & {338.65} & {944.92} \\\midrule
\textsc{Focal-R} + \textbf{\textsc{RankSim}}   
& {7.77} & {6.99} & 14.23 & 26.01 & {4.38} & {4.03} & 9.25 & 20.16  & {130.86} & \textbf{102.82} & {342.70} & 967.78 \\
\textsc{Focal-R} + \textsc{LDS} + \textbf{\textsc{RankSim}}   
& 7.71 & 6.99 & \textbf{13.65} & {24.97} & 4.31 & 3.98 & \textbf{8.72} & {17.56} & 131.63 & 104.88 & \textbf{333.44} & {932.95} \\
\textsc{Focal-R} + \textsc{FDS} + \textbf{\textsc{RankSim}} & {7.75} & 7.01 & {14.06} & \textbf{24.56} & {4.33} & {3.99} & {9.04} & \textbf{16.26}& {132.42} & {105.45} & {338.61} & \textbf{918.23}\\
\textsc{Focal-R} + \textsc{LDS} + \textsc{FDS} + \textbf{\textsc{RankSim}} & \textbf{7.67} & \textbf{6.91} & 14.07 & 25.01 & \textbf{4.28} & \textbf{3.93} & 9.38 & 18.41& \textbf{130.20} & {102.94} & 338.73 & {923.43}\\
\midrule\midrule
RRT      
& 7.81 & 7.07 & 14.06 & 25.13 & 4.35 & 4.03 & 8.91 & 16.96 & 132.99 & 105.73 & 341.36 & 928.26\\
RRT + LDS           
& 7.79 & 7.08 & 13.76 & 24.64 & 4.34 & {4.02} & 8.72 & 16.92 & 132.91 & 105.97 & 338.98 & 916.98\\
RRT + FDS           
& 7.65 &  {7.02} & 12.68 & 23.85 & 4.31 & 4.03 & 7.58 & 16.28 & 129.88 & {104.63} & 310.69 & 890.04\\
RRT + LDS + FDS     
&  {7.65} & 7.06 &  {12.41} & {23.51} & {4.31} & 4.07 & {7.17} & {15.44} & {129.14} & 105.92 & {306.69} & {880.13}\\\midrule
\textsc{RRT} +\textbf{\textsc{RankSim}}    
& 7.55 & 6.83 & 13.47 & 24.72 & 4.17 & {3.86} & 8.66 & 15.54 & 127.78 & 101.16 & 325.96 & 945.92 \\
\textsc{RRT} + \textsc{LDS} + \textbf{\textsc{RankSim}}   
& 7.56 & 6.83 & 13.06 & 24.78 & 4.23 & 3.91 & 8.55 & 17.44 & 127.13 & \textcolor{bestblue}{\textbf{100.05}} & 329.27 & 954.99 \\
\textsc{RRT} + \textsc{FDS} + \textbf{\textsc{RankSim}} 
&  \textcolor{bestblue}{\textbf{7.35}} & 6.81 & \textcolor{bestblue}{\textbf{11.50}} & \textbf{22.75} & \textcolor{bestblue}{\textbf{4.05}} & {3.85} & \textcolor{bestblue}{\textbf{6.05}} & \textbf{14.68} & \textcolor{bestblue}{\textbf{123.18}} & 100.86 & \textcolor{bestblue}{\textbf{280.55}} & \textbf{879.85}\\
\textsc{RRT} + \textsc{LDS} + \textsc{FDS} + \textbf{\textsc{RankSim}} &  {7.37} & \textcolor{bestblue}{\textbf{6.80}} &  {11.83} & {23.11} & 4.06 & \textcolor{bestblue}{\textbf{3.84}} & {6.33} & {14.71} & {123.61} & {100.64} & {287.16} & {889.62}\\
\midrule\midrule
\textsc{SQInv}  & 7.87 & 7.24 & 12.44 & 22.76 & 4.47 & 4.22 & 7.25 & 15.10 & 134.36 & 111.23 & 308.63 & 834.08\\
\textsc{SQInv} + \textsc{LDS}         & 7.83 & 7.31 & 12.43 & 22.51 & 4.42 & 4.19 & 7.00 & 13.94 & 131.65 & 109.04 & {\textcolor{black}{298.98}} & 829.35\\
\textsc{SQInv} + \textsc{FDS}           & 7.83 & 7.23 & 12.60 & 22.37 & 4.42 & 4.20 & 6.93 & 13.48& 132.64 & 109.28 & 311.35 & 851.06 \\
\textsc{SQInv} + \textsc{LDS}   + \textsc{FDS}     & 7.78 & 7.20 & 12.61 & {22.19} & 4.37 & 4.12 & 7.39 &  {12.61} & 129.35 & 106.52 & 311.49 & {811.82}  \\ 
\midrule
\textsc{SQInv} + \textbf{\textsc{RankSim}} 
& \textcolor{black}{\textbf{7.42}} & \textbf{6.84} & \textbf{12.12} & 22.13 & \textbf{4.10} & \textbf{3.87} & \textbf{6.74} &  {12.78} & \textbf{123.76} & \textbf{101.02} & \textbf{296.73} & {793.55}\\
\textsc{SQInv} + \textsc{LDS} +\textbf{\textsc{RankSim}}
& {7.57} & {7.00} & {12.16} & {22.44} & {4.23} & {4.00} & {6.81} & 13.23 & {127.29} & {104.07} & {299.81} & 844.00 \\
\textsc{SQInv} + \textsc{FDS}  +\textbf{\textsc{RankSim}}
& 7.50 & {6.93} & 12.09 & 21.68 & {4.19} & {3.97} & 6.65 & 13.28 & 125.30 & {102.68} & {299.10} & \textcolor{bestblue}{\textbf{777.48}}\\
\textsc{SQInv} + \textsc{LDS} + \textsc{FDS}  +\textbf{\textsc{RankSim}}  
& {7.69} & {7.13} & {12.30} & \textcolor{bestblue}{\textbf{21.43}} & {4.34} & 4.13 & {6.72} &  \textcolor{bestblue}{\textbf{12.48}} & {129.09} & {106.21} & {303.82} & {798.05}\\

\bottomrule[1.5pt]
\end{tabular}}
\end{center}
\vskip -0.1in
\end{table}

%% file: chapters/tables/supp_agedb.tex
\begin{table}[h]
\setlength{\tabcolsep}{2.5pt}
\caption{Complete Results on \textit{AgeDB-DIR}. The {best} results for each method (Vanilla, Focal-R, RRT, SQINV) are in \textbf{bold}. The best results for each metric and data subset (entire column) are in \textbf{\textcolor{bestblue}{bold and red}}. We use batch size of 64 and learning rate of 2.5e-4 for RankSim. Baseline numbers are quoted from \cite{yangetal2021}. We find in some experiments baselines can perform better in the setting of batch size of 64. Thus we include the results with batch size of 64 in brackets for fair comparison.}
\vspace{-4pt}
\label{table:agedb_supp}
\begin{center}\resizebox{1\textwidth}{!}{
\begin{tabular}{l|cccc|cccc|cccc}
\toprule[1.2pt]
\textbf{Metrics} & \multicolumn{4}{c|}{\textbf{MAE} $\bf{\downarrow}$} & \multicolumn{4}{c|}{\textbf{GM} $\bf{\downarrow}$} & \multicolumn{4}{c}{\textbf{MSE} $\downarrow$}\\\midrule
\textbf{Shot} & All & Many & Med. & Few & All & Many & Med. & Few& All & Many & Med. & Few\\
 \midrule\midrule
 \textsc{Vanilla}   
 & 7.77 (7.35) & 6.62 (6.56) & 9.55 (8.23) & 13.67 (12.37) 
 & 5.05 (4.82) & 4.23 (4.32) & 7.01 (5.49) & 10.75 (9.47) 
 & 101.60 (93.47) & 78.40 (73.27) & 138.52 (121.67) & 253.74 (241.97) \\  
\textsc{Vanilla} + \textsc{LDS} 
& 7.67 (7.39) & 6.98 (6.72) & 8.86 (8.18) & 10.89 (10.25) & 4.85 (4.65) & 4.39 (4.30) & 5.80 (5.25) & 7.45 (6.91) & 102.22 (91.79) & 83.62 (77.24) & 128.73 (111.52) & 204.64 (174.73)\\  
\textsc{Vanilla} + \textsc{FDS} & 7.55 (7.42) &  6.50 (6.46) & 8.97 (8.64) & 13.01 (13.07) & 4.75 (4.77) & 4.03 (4.13) & 6.42 (5.96) & 9.93 (10.27) & 98.55 (93.42) & 75.06 (70.29) & 123.58 (118.25) & 235.70 (243.12) \\  
\textsc{Vanilla} + \textsc{LDS} + \textsc{FDS} 
& 7.55 (7.44) & 7.01 (6.85) & 8.24 (8.36) & 10.79 (10.43) & 4.72 (4.79) & 4.36 (4.34) & 5.45 (5.79) & 6.79 (7.15) & 99.46 (92.38) & 84.10 (78.99) & 112.20 (107.32) & 209.27 (177.45) \\ \midrule 
\textsc{Vanilla} + \textbf{\textsc{RankSim}} & {7.13}  & {6.51}   & {8.17} & \textbf{10.12} & {4.48} & {4.01} & {5.27} & \textbf{\textcolor{black}{6.79}} & {87.45} & 71.84 & {111.41} & {168.61}  \\
\textsc{Vanilla} + \textsc{LDS} +  \textbf{\textsc{RankSim}} & \textbf{6.99} & \textbf{6.38} & {7.88} & {10.23} & \textbf{4.40} & \textbf{3.97} & {5.30} & {6.93} & \textbf{84.14} & {71.72} & \textbf{\textcolor{black}{98.59}} &  \textbf{{161.48}}   
\\
\textsc{Vanilla} + FDS +  \textbf{\textsc{RankSim}}    & 7.33 & {6.49} & 8.53 & 11.98 & 4.82 & 4.19 & 6.16 & 8.99 & 90.09 & \textbf{69.90} & 112.69 & 205.12 \\\color{black}
\textsc{Vanilla} + LDS + FDS +  \textbf{\textsc{RankSim}}    
& 7.03 & {6.54} & \textbf{7.68}  & \textbf{9.92} & 4.45 & {4.07} &  \textbf{5.23}  &  \textbf{6.35} & 84.96 & 74.27 & \textcolor{black}{\textbf{93.64}} & 161.92\\\midrule\midrule
\textsc{Focal-R} 
& 7.64 (7.41) & 6.68 (6.63) & 9.22 (8.54) & 13.00 (11.68) & 4.90 (4.72) & 4.26 (4.28) & 6.39 (5.36) & 9.52 (8.39) & 101.26 (94.08) & 77.03 (74.99) & 131.81 (116.39) & 252.47 (212.67) \\ 
\textsc{Focal-R} + {\textsc{LDS}} & 7.56 (7.25) & 6.67 (6.41) & 8.82 (8.48) & 12.40 (11.77) & 4.82 (4.68) & 4.27 (4.05) & 5.87 (5.74) & 8.83 (7.84) & 98.80 (90.15) & 77.14 (70.01) & 125.53 (112.42) & 229.36 (216.84)   \\ 
\textsc{Focal-R} + {\textsc{FDS}} & 7.65 (7.33) & 6.89 (6.51) & 8.70 (8.49) & 11.92 (11.97) & 4.83 (4.78) & 4.32 (4.23) & 5.89 (5.79) & 8.04 (8.80) & 100.41 (90.73) & 80.97 (71.64) & 121.84 (113.45) & 208.25 (208.17)    \\
\textsc{Focal-R} + {\textsc{LDS}} + {\textsc{FDS}} 
& 7.47 (7.32) & 6.69 (6.60) & 8.99 ({8.26}) & 12.83 (11.53) & 4.93 (4.75) & 4.27 (4.26) & 6.25 (5.61) & 9.98 (8.45) & 96.51 (90.91) & 75.02 (74.23) & 123.43 ({108.77}) & 225.04 (199.10)  \\ \midrule
\textsc{Focal-R} + \textbf{\textsc{RankSim}}   & {7.15} & {6.45} & {7.97}  & {11.50} & \textbf{4.53} & {4.10} & {5.10}  & {8.50} & \textbf{86.67} & {69.57} & \textbf{105.10} & {197.15}\\ 
\textsc{Focal-R} + \textsc{LDS} + \textbf{\textsc{RankSim}}   & 7.25 & {6.40} & 8.71 & {11.24} & {4.58} & {4.02} & 5.99 & \textbf{7.52} & 89.54 & 69.87 & 118.84 & \textbf{194.34} \\
\textsc{Focal-R} + FDS + \textbf{\textsc{RankSim}}   & 7.25 & 6.72 & \textbf{7.86} & \textbf{10.58} & 4.54 & 4.22 &\textbf{\textcolor{bestblue}{4.84}} & {7.57} & 90.12 & 76.69 & {105.55} & 175.18\\ \color{black}
\textsc{Focal-R} + LDS + FDS +\textbf{\textsc{RankSim}}   &\textbf{7.09} & \textbf{\textcolor{bestblue}{6.17}} & 8.71 & 11.68 & 4.46 & \textbf{\textcolor{bestblue}{3.85}} & 5.76 & 8.78 & {87.12} & \textbf{\textcolor{bestblue}{66.01}} & 120.10 & {195.35} \\ 
\midrule\midrule
\textsc{RRT} 
& 7.74 (7.29) & 6.98 (6.68) & 8.79 (7.89) & 11.99 (11.36) & 5.00 ({4.61}) & 4.50 ({4.24}) & 5.88 ({5.01}) & 8.63 (8.13) & 102.89 (91.30) & 83.37 (75.21) & 125.66 (108.59) & 224.04 (195.38) \\
\textsc{RRT} + \textsc{LDS} 
& 7.72 (7.52) & 7.00 (6.86) & 8.75 (8.22) & 11.62 (11.82) & 4.98 (4.90) & 4.54 (4.42) & 5.71 (5.66) & 8.27 (8.63) & 102.63 (95.85) & 83.93 (79.85) & 126.01 (110.95) & 214.66 (205.24) \\
\textsc{RRT} + \textsc{FDS} 
& 7.70 (7.30) & 6.95 (6.64) & 8.76 (8.01) & 11.86 (11.44) & 4.82 (4.62) & 4.32 (4.24) & 5.83 (5.06) & 8.08 (8.09) & 102.09 (91.60) & 84.49 (74.79) & 122.89 (110.78) & 224.05 (197.30) \\
\textsc{RRT} + \textsc{LDS} + \textsc{FDS} 
& 7.66 (7.29) & 6.99 (6.70) & 8.60 ({7.85}) & 11.32 ({11.26}) & 4.80 (4.62) & 4.42 (4.28) & 5.53 (\textbf{4.95}) & 6.99 ({7.75}) & 101.74 (91.35) & 83.12 (75.93) & 121.08 ({107.46}) & 210.78 (192.41) \\\midrule
\color{black}
\textsc{RRT} + \textbf{\textsc{RankSim}} & {7.11} & {6.53} & {8.00}   & {10.04} & {4.52} & {4.19} & {5.05}  & \textbf{\textcolor{black}{6.77}} & {86.93} & \textbf{72.11} & {108.08} & {168.41} \\ 
 \textsc{RRT} + \textsc{LDS} + \textbf{\textsc{RankSim}}    & \textbf{6.94} & \textbf{6.43} & \textbf{\textcolor{bestblue}{7.54}} & {10.10} & \textbf{4.37} & \textbf{3.97} & 5.11 & {7.05} & \textbf{82.98} & 72.49 &\textbf{\textcolor{bestblue}{91.24}} & \textbf{159.24}\\
 \color{black}
  \textsc{RRT} + \textsc{FDS} + \textbf{\textsc{RankSim}}  & {7.11} & 6.55 & 7.99 & \textbf{10.02} & {4.49} & {4.13} & 5.13 & {6.85} & {86.93} & {72.36} & 107.90 & {166.55} \\
  \color{black}
 \textsc{RRT} + \textsc{LDS} + \textsc{FDS} + \textbf{\textsc{RankSim}}  
 & 7.13 & 6.54 & 8.07 & 10.12 & 4.55 & 4.18 & 5.20 & 6.87 & 87.28 & {72.20} & 109.19 & 169.16\\
 \midrule\midrule
\textsc{SQInv} 
& 7.81 (7.49) & 7.16 (6.92) & 8.80 (8.34) & 11.20 (10.55) & 4.99 (4.83) & 4.57 (4.53) & 5.73 ({5.25}) & 7.77 (7.05) & 105.14 (93.89) & 87.21 (77.68) & 127.66 (117.02) & 212.30 (183.03)  \\
\textsc{SQInv} + {\textsc{LDS}}  
& 7.67 (7.42) & 6.98 (6.83) & 8.86 (8.21) & 10.89 (10.79) & 4.85 (4.73) & 4.39 (4.37) & 5.80 (5.40) & 7.45 (7.03) & 102.22 (95.10) & 83.62 (78.63) & 128.73 (111.75) & 204.64 (204.60)\\ 
\textsc{SQInv} + {\textsc{FDS}} 
& 7.69 (7.55) & 7.10 (6.99) & 8.86 (8.40) & 10.89 (10.48) & 4.85 (4.82) & 4.39 (4.49) & 5.80 (5.47) & 7.45 (6.58) & 102.22 (97.35) & 83.62 (81.90) & 128.73 (118.30) & 204.64 ({155.32}) \\  
\textsc{SQInv} + {\textsc{LDS}} + {\textsc{FDS}} 
& 7.55 (7.37) & 7.01 (6.82) & 8.24 (8.25) & 10.79 ({10.16}) & 4.72 (4.72) & 4.36 (4.33) & 5.45 (5.50) & 6.79 (6.98) & 99.46 (93.45) & 84.10 (79.86) & 112.20 (110.00) & 209.27 (184.41)\\\midrule
\textsc{SQInv} + \textbf{\textsc{RankSim}}    & \textbf{\textcolor{bestblue}{6.91}} & \textbf{\textcolor{black}{6.34}} & {7.79}   & {9.89} & \textbf{\textcolor{bestblue}{4.28}} & \textbf{\textcolor{black}{3.92}} & \textbf{\textcolor{black}{4.88}} & {6.89}& \textbf{\textcolor{bestblue}{82.10}} & \textbf{\textcolor{black}{68.60}} & {102.61} & {152.84}  \\
\textsc{SQInv} + {\textsc{LDS}} + \textbf{\textsc{RankSim}}     & {6.99} & {6.38} & {7.88} & 10.23 & {4.40} & {3.97} & 5.30 & {6.90} & {84.14} & {71.72} & {98.59} &  {{161.48}} \\
\textsc{SQInv} + {\textsc{FDS}} + \textbf{\textsc{RankSim}} & 7.02 & {6.49} & {7.84} & \textbf{\textcolor{bestblue}{9.68}}  & 4.53 & {4.13} &  5.37  & {6.89} & {83.51} & {71.99} & {99.14} & \textcolor{bestblue}{\textbf{149.05}} \\[1.2pt]
\color{black}
\textsc{SQInv} + {\textsc{LDS}} + {\textsc{FDS}} + \textbf{\textsc{RankSim}} & 7.03 & {6.54} & \textbf{7.68}  & 9.92 & 4.45 & {4.07} & {5.23}  & \textcolor{bestblue}{\textbf{6.35}}  & 84.96 & 74.27 & \textcolor{black}{\textbf{93.64}} & 161.92 \\
\bottomrule[1.2pt]
\end{tabular}}
\end{center}
\vspace{-0.5cm}
\end{table}

%% file: chapters/tables/supp_sts.tex
\begin{table}[h]
\setlength{\tabcolsep}{2.5pt}
\caption{Complete Results on \textit{STS-B-DIR}. The {best} results for each method (Vanilla, Focal-R, RRT, SQINV) are in \textbf{bold}. The best results for each metric and data subset (entire column) are in \textbf{\textcolor{bestblue}{bold and red}}. For RankSim, we use batch size of 16 and learning rate of 2.5e-4. Baseline numbers are quoted from \cite{yangetal2021}.}\label{table:sts_supp}
\small
\begin{center}
\resizebox{0.9\textwidth}{!}{
\begin{tabular}{l|cccc|cccc|cccc|cccc}
\toprule[1.5pt]
\textbf{Metrics}      & \multicolumn{4}{c|}{\textbf{MSE}~$\downarrow$} & \multicolumn{4}{c|}{\textbf{MAE}~$\downarrow$}      & \multicolumn{4}{c|}{\textbf{Pearson correlation~(\%)}~$\uparrow$} &\multicolumn{4}{c}{\textbf{Spearman correlation~(\%)}~$\uparrow$}    \\ \midrule
Shot         & All   & Many  & Med. & Few   & All& Many  & Med. & Few & All & Many & Med. & Few & All & Many & Med. & Few   \\ \midrule\midrule
\textsc{Vanilla} 
& 0.974 & 0.851 & 1.520  & 0.984 & 0.794 & 0.740 & 1.043 & 0.771 & 74.2 & 72.0 & 62.7  & 75.2 & 74.4 & 68.8 & 50.5 & 75.0\\
\textsc{Vanilla} + LDS 
& 0.914 & {0.819} 
& 1.319 & 0.955 & 0.773 & {0.729} & 0.970 & 0.772  & 75.6 & {73.4} & 63.8 & 76.2 & 76.1 &  {70.4} & \textcolor{bestblue}{\textbf{55.6}} & 74.3 \\
\textsc{Vanilla} + {FDS} 
& 0.916 & 0.875 & {1.027} & 1.086 & 0.767 & 0.746 & {0.840} & 0.811  & 75.5 & 73.0 & {67.0} & 72.8 & 75.8 & 69.9 & 54.4 & 72.0 \\
\textsc{Vanilla} + LDS + FDS 
& {0.907} & \textcolor{bestblue}{\textbf{0.802}} & 1.363  & {0.942}  & {0.766} & \textcolor{bestblue}{\textbf{0.718}} & 0.986 & {0.755}   & {76.0}     & \textcolor{bestblue}{\textbf{74.0}} & 65.2      & {76.6} & {76.4} & \textbf{70.7} & {54.9} & 74.9  \\
\midrule
\textsc{Vanilla} + \textbf{\textsc{RankSim}}
& \textcolor{black}{\textbf{0.873}} & 0.908 & \textcolor{bestblue}{\textbf{0.767}}  & 0.705 & \textcolor{black}{\textbf{0.749}} & {0.755} &  {0.737} & 0.695 & \textbf{76.8} & 71.0 & \textcolor{bestblue}{\textbf{72.9}}  & {85.2} & \textcolor{black}{\textbf{77.2}} & 68.3 & 55.4 & \textbf{88.4} \\
\textsc{Vanilla} + \textsc{LDS} +  \textbf{\textsc{RankSim}} & 0.889 & 0.911 & 0.849 & 0.690  & 0.755 & 0.762 & 0.758 & \textcolor{bestblue}{\textbf{0.638}}  & 76.2 & 70.7 & 70.0 & \textbf{85.6} & 76.3 & 67.8 & 49.0 & 85.4 \\
\textsc{Vanilla} + \textsc{{FDS}}  + \textbf{\textsc{RankSim}}  
& 0.884 & 0.924 & \textcolor{bestblue}{\textbf{0.767}} &  \textbf{0.685}  & 0.755 & 0.769 & \textbf{0.736} & 0.653 & 76.5 & 70.4 & 72.5 & 85.7 & 76.7 & 67.1 & 53.5 & 87.8   \\ 
\textsc{Vanilla} + \textsc{{LDS}} + \textsc{{FDS}} +  \textbf{\textsc{RankSim}}  & 0.903 & 0.908 & 0.911 & 0.804  & 0.761 & 0.759 & 0.786 & 0.712   & 75.8 & 70.6 & 69.0 & 82.7 & 75.8 & 67.3 & 49.3 & 84.9  \\ 
\midrule\midrule
\textsc{Focal-R}      
& 0.951 & {0.843} & 1.425  & 0.957 & 0.790 & 0.739 & 1.028 & 0.759 & 74.6 & 72.3 & 61.8  & 76.4 & 75.0 & 69.4 & 51.9 & 75.5\\
\textsc{Focal-R} + LDS 
& 0.930     & \textbf{0.807}     & 1.449      & 0.993    & 0.781 & \textbf{0.723} & 1.031 & {0.801} & {75.7}     & \textbf{73.9}     & 62.4      & 75.4 & {76.2} & \textcolor{bestblue}{\textbf{71.2}} & 50.7 & 74.7\\
\textsc{Focal-R} + {{FDS}} 
& {0.920} & 0.855 & {1.169}  & 1.008     & {0.775}& 0.743 & {0.903}&0.804 & {75.1} & {72.6}   & {66.4} & 74.7 & 75.4 & {69.4}&\textbf{52.7}&75.4\\
\textsc{Focal-R} + {{LDS}} + {{FDS}} 
& 0.940 & 0.849 & 1.358  & {0.916}  & {0.785}& {0.737}& 0.984& 0.732& 74.9 & 72.2 & 66.3 & {77.3} & 75.1 & 69.2 & {52.5} & {76.4}\\ \midrule
\textsc{Focal-R} +  \textbf{\textsc{RankSim}} 
& {0.887} & 0.889 & {0.918}  & {0.745}    & {0.763} &{0.757} & {0.805}& {0.719} & {76.2}     & 70.8     & {70.4}      & {84.6} & {76.7} & 68.2 & 51.0 & 87.7 \\
% \textsc{Focal-R} + \textsc{Inv} +  \textbf{\textsc{RankSim}} & 0.909 & 0.920 & 0.919  & \textbf{0.694}  & {0.764} & {0.762} & {0.791} & {0.704} & 75.7 & 69.9 & 68.2 & 85.5 & 76.0 & 66.9 & 49.7 & 88.9\\
\textsc{Focal-R} + \textsc{LDS}  + \textbf{\textsc{RankSim}}
& \textbf{0.872} & 0.887 & {0.847} & \textbf{0.718} & \textbf{0.752} & 0.751 & {0.770} & {0.701} & \textbf{76.7} & 71.2 & 70.3 & \textbf{85.1} & \textbf{77.1} & 68.2 & 50.1 & \textbf{88.4} \\
\textsc{Focal-R} + \textsc{FDS}  + \textbf{\textsc{RankSim}} 
& 0.913 & 0.952 & 0.793 & 0.723  & 0.763 & 0.776 & 0.735 & \textbf{0.660} & 75.6 & 69.6 & \textbf{71.5} & 84.8 & 75.8 & 66.9 & 49.7 & 87.7  \\
\textsc{Focal-R} + \textsc{LDS} + \textsc{FDS} + \textbf{\textsc{RankSim}} & 0.911 & 0.943 & \textbf{0.779} & 0.866  & 0.757 & 0.765 & \textcolor{bestblue}{\textbf{0.725}} & 0.747 & 75.7 & 69.9 & 71.4 & 81.2 & 75.7 & 67.0 & 47.7 & 80.7\\
\midrule\midrule
\textsc{RRT} 
& 0.964 & 0.842 & 1.503  & 0.978 & 0.793 & 0.739 & 1.044 & 0.768 & 74.5 & 72.4 & 62.3  & 75.4 & 74.7 & 69.2 & 51.3 & {74.7}
 \\
\textsc{RRT} + LDS 
& {0.916} & {0.817} & {1.344}  & 0.945 & 0.772 & {0.727} & 0.980 & {0.756} & {75.7} & {73.5} & {64.1}  & {76.6} & 76.1 & {70.4} & {53.2} & 74.2\\
\textsc{RRT} + FDS 
& 0.929 & 0.857 & {1.209}  & 1.025  & 0.769 & 0.736 & {0.905} & 0.795   & 74.9     & 72.1     & {67.2}      & 74.0 & 75.0 & 69.1 & 52.8 & 74.6\\
\textsc{RRT} + LDS + FDS 
& {0.903} & \textbf{0.806} & 1.323  & {0.936}  & {0.764} & \textbf{0.719} & 0.965 & 0.760   &  {76.0}     &\textbf{73.8}     & 65.2      & {76.7}  & {76.4} & \textbf{70.8} & \textbf{54.7} & {74.7}\\\midrule
\textsc{RRT} + \textbf{\textsc{RankSim}} 
% & 0.887 & 0.905 & 0.861  & 0.706  & 0.758 & 0.760 & 0.768 & 0.684 &  76.5 & 71.7 & 68.0 & 85.2  & 76.9 & 69.1 & 48.4 & 88.2 \\ 
%  
% \textsc{RRT} + \textsc{Inv} + \textbf{\textsc{RankSim}}
& \textcolor{bestblue}{\textbf{0.865}} & 0.876 & 0.867  & \textcolor{bestblue}{\textbf{0.670}}  & \textcolor{bestblue}{\textbf{0.748}} & 0.749 & 0.767 & \textbf{0.670}  &  \textcolor{bestblue}{\textbf{77.1}} & 72.2 & 68.3 & \textcolor{bestblue}{\textbf{86.1}}  & \textcolor{bestblue}{\textbf{77.4}} & 69.6 & 48.0 & \textcolor{bestblue}{\textbf{89.4}} \\
\textsc{RRT} + \textsc{LDS} + \textbf{\textsc{RankSim}} 
& 0.874 & 0.893 & \textbf{0.833}  & 0.722  & 0.754 & 0.758 & \textbf{0.752} & 0.698   &  77.0 & 72.3 & 68.3 & 84.8  & 77.2 & 69.5 & 47.2 & 87.8\\
\textsc{RRT} + \textsc{FDS} + \textbf{\textsc{RankSim}} & {0.871} & 0.874 & 0.898  & 0.734  & 0.750 & 0.748 & 0.779 & 0.687   &  76.8 & 72.0 & \textbf{68.7} & 84.5  & 77.0 & 69.4 & 47.1 & 87.2 \\
\textsc{RRT} + \textsc{LDS} + \textsc{FDS} + \textbf{\textsc{RankSim}} 
& 0.882 & 0.892 & 0.887 & 0.702 & 0.758 & 0.759 & 0.775 & 0.681 & 76.6 & 71.7 & 68.0 & 85.5 & 76.8 & 69.0 & 46.5 & 88.3\\
\midrule
% \textsc{RRT} + \textbf{\textsc{RankSim}}   & \textbf{0.883} & 0.905 & \textbf{0.827} & \textbf{0.744} & \textbf{0.759} & 0.766 & \textbf{0.750} & \textbf{0.691} & \textbf{76.5} & 71.1 & \textbf{69.8} & \textbf{84.3} & \textbf{76.9} & 68.7 & 50.4 & \textbf{86.4}  \\ \midrule\midrule
\textsc{Inv} 
& 1.005 & 0.894 & 1.482  & 1.046 & 0.805 & 0.761 & 1.016 & 0.780 & 72.8 & 70.3 & 62.5  & 73.2 & 73.1 & 67.2 & 54.1  & 71.4 \\
\textsc{Inv} + {{LDS}} 
& 0.914 & {0.819} & {1.319}  & {0.955} & 0.773 & {0.729} & 0.970 & 0.772 & {75.6} & {73.4} & {63.8}  & {76.2} & 76.1 & {70.4} & \textcolor{bestblue}{\textbf{55.6}}  & 74.3\\
\textsc{Inv} + FDS 
& 0.927 & 0.851 & {1.225}  & 1.012   & 0.771 & 0.740 & {0.914} & 0.756  & 75.0     & 72.4     & {66.6} & 74.2  & 75.2 & 69.2 & {55.2} & 74.8\\
\textsc{Inv} + LDS + FDS 
& {0.907} & \textcolor{bestblue}{\textbf{0.802}} & 1.363  & {0.942}  & {0.766} & \textcolor{bestblue}{\textbf{0.718}} & 0.986 & {0.755}   & {76.0}     & \textcolor{bestblue}{\textbf{74.0}}     & 65.2      & {76.6} & \textbf{76.4} & \textbf{70.7} & 54.9 & {74.9}\\\midrule
\textsc{Inv} + \textbf{\textsc{RankSim}}  & 1.091 & 1.056 & 1.240 & 1.118 & 0.854 & 0.843 & 0.912 & 0.822 & 69.9 & 65.2 & 60.1 & 76.0 & 70.2 & 62.5 & 45.0 & 78.5 \\
\textsc{Inv} + \textsc{LDS}  + \textbf{\textsc{RankSim}} & \textbf{0.889} & 0.911 & \textbf{0.849} & \textbf{0.690} & \textbf{0.755} & 0.762 & \textbf{0.758} & \textcolor{bestblue}{\textbf{0.638}}  & \textcolor{black}{\textbf{76.2}} & 70.7 & \textbf{70.0} & \textbf{85.6} & 76.3 & 67.8 & 49.0 & \textbf{85.4}  \\
\textsc{Inv} + \textsc{FDS}  + \textbf{\textsc{RankSim}}  & 1.083 & 1.035 & 1.301 & 1.063 & 0.831 & 0.812 & 0.914 & 0.840  & 70.0 & 64.8 & 68.9 & 76.7 & 69.7 & 61.6 & 43.8 & 82.5 \\
\textsc{Inv} + \textsc{LDS} + \textsc{FDS} + \textbf{\textsc{RankSim}} & 0.903 & 0.908 & 0.911 & 0.804  & 0.761 & 0.759 & 0.786 & 0.712   & 75.8 & 70.6 & 69.0 & 82.7 & 75.8 & 67.3 & 49.3 & 84.9  \\
\bottomrule[1.5pt]
\end{tabular}
}
\end{center}
\vspace{-0.5cm}
\end{table}

%% file: chapters/tables/hyperparameter.tex
\begin{table}[h]
\setlength{\tabcolsep}{2.5pt}
\caption{Varying hyperparameters $\gamma$ and $\lambda$. We conduct experiments on \textit{AgeDB-DIR} with SQINV. In the first section, we keep $\lambda$ as 2 and change the value of $\gamma$ (the best result is in \textbf{bold}). In the second section, we keep $\gamma$ as 100 and change the value of $\lambda$ (the best result is in \textbf{bold}). The best result for each metric in all experiments (entire column) is in \textbf{\textcolor{bestblue}{bold and red}}. }
\vspace{-4pt}
\label{table:hyperparams_agedb}
\small
\vskip 0.15in
\begin{center}
\resizebox{0.62\linewidth}{!}{
\begin{tabular}{lc|cccc|cccc|cccc}
\toprule[1.5pt]
\multicolumn{2}{c|}{\textbf{hyperparams}}  & \multicolumn{4}{c|}{\textbf{MAE~$\downarrow$}}       & \multicolumn{4}{c|}{\textbf{GM~$\downarrow$}}  & \multicolumn{4}{c}{\textbf{MSE~$\downarrow$}}  \\ 
$\gamma$  & $\lambda$ & All   & Many  & Med. & Few   & All   & Many   & Med. & Few & All   & Many  & Med. & Few   \\ \midrule%[1.5pt]
% w/o SimRank  
% & 7.49 & 6.92 & 8.34 & 10.55 & 4.83 & 4.53 & {5.25} & 7.05 & 93.89 & 77.68 & 117.02 & 183.03  \\
%  \midrule\midrule
\textbf{0.01} & 2  
& 7.42 & 6.75 & 8.42 & 11.01 & 4.76 & 4.31 & 5.63 &  7.59 & 94.68 & 77.19 & 114.39 & 205.43 \\
\textbf{0.1} & 2 
& 7.36 & 6.72 & 8.44  & 10.42 & 4.72 & 4.31 & 5.58 &  7.00 & 91.86 & 76.50 & 114.43 & 174.48 \\
\textbf{1} & 2 
& 7.17 & 6.56 & 8.23 & 10.08 & 4.49 & 4.10 & 5.30 & 6.78 & 89.20 & 75.28 & 109.53 & 164.49\\
\textbf{10} & 2 
& 7.05 & 6.55 & \textbf{7.79} & {9.72} & 4.56 & 4.22 & 5.18 & {6.53} & 84.57 & 72.71 & \textcolor{black}{\textbf{99.37}} & 155.74  \\
\textbf{100} & 2  
& \textcolor{bestblue}{\textbf{6.91}} &  \textcolor{bestblue}{\textbf{6.34}} & \textbf{7.79}   & {9.89} & \textcolor{bestblue}{\textbf{4.28}} & \textcolor{bestblue}{\textbf{3.92}} & \textcolor{bestblue}{\textbf{4.88}} & {6.89}& \textcolor{bestblue}{\textbf{82.10}} & \textcolor{bestblue}{\textbf{68.60}} & {102.61} & {152.84}\\
\textbf{200} & 2  
& 7.15 & 6.52 & 8.31 & 9.94 & 4.54 & 4.10 & 5.67 & \textbf{6.45} & 87.45 & 73.51 & 108.96 & 159.55  \\
\textbf{500} & 2  
& 7.18 & 6.54 & 8.17 & 10.46 & 4.55 & 4.12 & 5.45 & 7.01 & 88.86 & 73.57 & 110.19 & 174.29 \\
\textbf{1000} & 2  
& 7.17 & 6.47 & 8.44 & 10.26 & 4.54 & 4.13 & 5.46 & 6.80 & 89.71 & 72.37 & 119.44 & 171.34 \\
\textbf{2000} & 2  
& 7.10 & 6.50 & 7.99 & 10.27 & 4.50 & 4.08 & 5.35 & 7.06 & 87.23 & 72.98 & 103.97 & 175.56\\
\textbf{3000} & 2  
& 7.14 & 6.70 & 7.73 & \textcolor{black}{\textbf{9.66}} & 4.61 & 4.33 & 4.99 & 6.72 & 86.76 & 75.35 & 102.49 & \textcolor{bestblue}{\textbf{151.02}}\\
\midrule
100 & \textbf{0.05}  
& 7.76 & 7.02 & 9.25 & 10.67 & 4.97 & 4.53 & 6.06 & 6.92 & 103.20 & 83.02 & 140.37 & 191.01\\
100 & \textbf{0.1}  
& 7.32 & 6.72 & 8.21 & 10.55 & 4.65 & 4.25 & 5.47 & 6.98 & 91.37 & 76.59 & 110.95 & 176.76\\
100 & \textbf{0.5}  
& 7.11 & 6.64 & \textcolor{bestblue}{\textbf{7.70}} & 9.93 & 4.60 & 4.28 & 5.09 & 6.96 & 86.61 & 74.46 & {99.88} & 164.67 \\
100 & \textbf{1} 
& 7.26 & 6.63 & 8.31 & 10.24 & 4.63 & 4.18 & 5.64 & 6.93 & 90.42 & 76.13 & 113.52 & 161.46\\
100 & \textbf{1.5}  
& 7.07 & 6.45 & 8.22 & 9.79 & 4.47 & 4.06 & 5.40 & 6.66 & 85.11 & 71.15 & 107.27 & 155.61\\
100 & \textbf{2}  
&  \textcolor{bestblue}{\textbf{6.91}} &  \textcolor{bestblue}{\textbf{6.34}} & {7.79}   & {9.89} & \textcolor{bestblue}{\textbf{4.28}} & \textcolor{bestblue}{\textbf{3.92}} & \textcolor{bestblue}{\textbf{4.88}} & {6.89}& \textcolor{bestblue}{\textbf{82.10}} & \textcolor{bestblue}{\textbf{68.60}} & {102.61} & {152.84}\\
100 & \textbf{4}
& 7.14 & 6.60 & 7.84 & 10.31 & 4.55 & 4.25 & 5.03 & 6.69 & 87.31 & 73.84 & 102.94 & 171.33\\
100 & \textbf{8}  
& 7.26 & 6.85 & 7.82 & \textcolor{bestblue}{\textbf{9.61}} & 4.68 & 4.42 & 5.10 & \textcolor{bestblue}{\textbf{6.36}} & 89.91 & 79.06 & 104.37 & \textcolor{black}{\textbf{152.39}}\\
100 & \textbf{16}  
& 7.14 & 6.66 & 7.76 & 9.91 & 4.56 & 4.27 & 4.99 & 6.56 & 87.02 & 74.28 & 104.00 & 160.32\\
100 & \textbf{32}  
& 7.25 & 6.64 & 8.28 & 10.18 & 4.78 & 4.35 & 5.77 & 7.02 & 87.77 & 73.31 & 107.10 & 170.77\\
100 & \textbf{64}  
& 7.11 & 6.51 & 7.90 & 10.69 & 4.55 & 4.10 & 5.47 & 7.19 & 85.06 & 71.60 & \textcolor{bestblue}{\textbf{96.55}} & 180.44\\
\bottomrule
\end{tabular}
}
\end{center}
\vspace{-0.5cm}
\end{table}
% \begin{table}
% \centering
% \setlength{\tabcolsep}{2.5pt}
% \caption{Different hyperparameters $\gamma$ and $\lambda$ of RankSim. We conduct experiments on \textit{AgeDB-DIR} with SQINV and MAE loss (batch size: 64, learning rate: 2.5e-4).}
% \vspace{10pt}
% \label{table:hyperparams}
% \scalebox{0.8}{
% % \resizebox{0.51\textwidth}{!}{
% \begin{tabular}{lcccccccccccccccc}
% \toprule
%   \multicolumn{2}{c}{\textbf{hyperparams}}&
%   &\multicolumn{4}{c}{\textbf{MAE~$\downarrow$}}& &\multicolumn{4}{c}{\textbf{GM~$\downarrow$}}&&\multicolumn{4}{c}{\textbf{MSE~$\downarrow$}}  \\
% \cmidrule{1-2}\cmidrule{4-7}\cmidrule{9-12}\cmidrule{14-17}
%   $\gamma$ & $\lambda$ & & All   & Many  & Med. & Few  && All  & Many  & Med. & Few  && All   & Many  & Med. & Few \\
% \midrule
% 100.0 & 2.0 &&  
% \textbf{6.91} & \textbf{6.34} & {7.79} & {9.89} &
% & \textbf{4.28} & \textbf{3.92} & 4.88 & {6.89} & 
% & \textbf{82.09} & \textbf{68.60} & {102.61} & {152.84}\\
% \bottomrule
% \end{tabular}}
% \vspace{-1em}
% \end{table}
% \caption{Different hyperparameters $\gamma$ and $\lambda$ of RankSim. We conduct experiments on \textit{AgeDB-DIR} with SQINV and MAE loss (batch size: 64, learning rate: 2.5e-4).}

\begin{table}[h]
\setlength{\tabcolsep}{2.5pt}
\caption{Varying hyperparameters $\gamma$ and $\lambda$. We conduct experiments on \textit{IMDB-WIKI-DIR} with SQINV. In the first section, we keep $\lambda$ as 2 and change the value of $\gamma$ (the best result is in \textbf{bold}). In the second section, we keep $\gamma$ as 100 and change the value of $\lambda$ (the best result is in \textbf{bold}). The best result for each metric in all experiments (entire column) is in \textbf{\textcolor{bestblue}{bold and red}}. }
\vspace{-4pt}
\label{table:hyperparams_imdb}
\small
\vskip 0.15in
\begin{center}
\resizebox{0.62\linewidth}{!}{
\begin{tabular}{lc|cccc|cccc|cccc}
\toprule[1.5pt]
\multicolumn{2}{c|}{\textbf{hyperparams}}  & \multicolumn{4}{c|}{\textbf{MAE~$\downarrow$}}       & \multicolumn{4}{c|}{\textbf{GM~$\downarrow$}}  & \multicolumn{4}{c}{\textbf{MSE~$\downarrow$}}  \\ 
$\gamma$  & $\lambda$ & All   & Many  & Med. & Few   & All   & Many   & Med. & Few & All   & Many  & Med. & Few   \\ \midrule%[1.5pt]
% w/o SimRank  
% & 7.49 & 6.92 & 8.34 & 10.55 & 4.83 & 4.53 & {5.25} & 7.05 & 93.89 & 77.68 & 117.02 & 183.03  \\
%  \midrule\midrule
\textbf{0.01} & 2  
& 7.67 & 7.08 & 12.42 & 22.86 & 4.32 & 4.07 & 7.25 & 13.52 & 129.26 & 105.88 & 302.88 & 851.47 \\
\textbf{0.1} & 2 
& 7.61 & 6.99 & 12.67 & 23.31 & 4.26 & 4.01 & 7.19 & 13.34 & 127.55 & 102.95 & 311.63 & 876.28 \\
\textbf{1} & 2 
& 7.47 & 6.88 & 12.25 & 22.89 & 4.17 & 3.93 & 7.02 & 14.08 & 125.15 & 101.97 & 295.09 & 859.42 \\
\textbf{10} & 2  
& 7.43 & \textcolor{black}{\textbf{6.83}} & 12.22 & 23.19 & 4.16 & 3.93 & 6.80 & 14.50 & 123.81 & 99.98 & 300.47 & 862.53 \\
\textbf{100} & 2 
& \textcolor{black}{{7.42}} & {6.84} & {12.12} & 22.13 & \textcolor{black}{\textbf{4.10}} & \textcolor{black}{\textbf{3.87}} & {6.74} &  \textcolor{black}{\textbf{12.78}} & {123.76} & {101.02} & {296.73} & {793.55}\\
\textbf{200} & 2 
& 7.44 & 6.88 & \textcolor{black}{\textbf{11.77}} & 22.75 & 4.13 & 3.91 & 6.53 & 14.93 & 124.68 & 102.11 & \textcolor{black}{\textbf{290.17}} & 838.68 \\
\textbf{500} & 2  
& 7.43 & 6.87 & 11.96 & 21.76 & 4.14 & 3.92 & 6.61 & 12.89 & 123.78 & 101.25 & 295.44 & 785.79 \\
\textbf{1000} & 2  
& \textcolor{black}{\textbf{7.41}} & \textcolor{black}{\textbf{6.83}} & 12.13 & \textcolor{black}{\textbf{21.54}} & 4.14 & 3.92 & 6.78 & 13.16 & \textcolor{bestblue}{\textbf{122.67}} & \textcolor{bestblue}{\textbf{99.85}} & 299.33 & \textcolor{bestblue}{\textbf{769.95}} \\
\textbf{2000} & 2  
& 7.58 & 6.98 & 12.35 & 22.81 & 4.22 & 3.98 & 6.89 & 15.13 & 127.88 & 103.97 & 306.62 & 856.13\\
\textbf{3000} & 2  
& 7.48 & 6.91 & 11.95 & 22.65 & 4.15 & 3.94 & \textcolor{bestblue}{\textbf{6.35}} & 14.14 & 126.39 & 103.21 & 298.52 & 842.27 \\
\midrule
100 & \textbf{0.05}  
& 7.41 & 6.87 & \textcolor{bestblue}{\textbf{11.72}} & 22.22 & 4.15 & 3.92 & 6.67 & 14.57 & 124.49 & 102.49 &  \textcolor{bestblue}{\textbf{284.39}} & 831.26 \\
100 & \textbf{0.1}  
& 7.41 & 6.84 & 11.89 & 22.34 & 4.11 & 3.89 & 6.60 & 13.32 & 124.32 & 101.87 & 289.18 & 833.23\\
100 & \textbf{0.5}   
& \textcolor{bestblue}{\textbf{7.37}} & \textcolor{bestblue}{\textbf{6.82}} & 11.76 & 22.05 & \textcolor{bestblue}{\textbf{4.05}} &  \textcolor{bestblue}{\textbf{3.83}} & \textbf{6.43} & 12.71 & 124.36 & 101.88 & 289.72 & 831.52\\
100 & \textbf{1} 
& 7.50 & 6.94 & 11.98 & 22.49 & 4.23 & 4.00 & 6.78 & 14.19 & 124.39 & 102.01 & 287.93 & 836.41\\
100 & \textbf{1.5}  
& 7.50 & 6.95 & 11.88 & 22.39 & 4.21 & 3.98 & 6.85 & 14.33 & 126.34 & 104.34 & 287.73 & 821.28\\
100 & \textbf{2}  
& 7.42 & {6.84} & {12.12} & 22.13 & \textbf{4.10} & {3.87} & {6.74} &  {12.78} & \textbf{123.76} & \textbf{101.02} & {296.73} & \textbf{793.55}\\
100 & \textbf{4} 
& 7.60 & 7.03 & 12.19 & 22.78 & 4.28 & 4.06 & 6.83 & 14.45 & 128.59 & 105.64 & 297.42 & 850.14\\
100 & \textbf{8} 
& 7.53 & 6.93 & 12.33 & 23.23 & 4.24 & 4.00 & 7.09 & 15.37 & 125.32 & 101.61 & 300.69 & 863.02\\
100 & \textbf{16}  
& 7.53 & 6.93 & 12.38 & 23.23 & 4.22 & 3.97 & 7.02 & 15.02 & 126.81 & 103.04 & 302.43 & 867.96\\
100 & \textbf{32} 
& 7.62 & 7.09 & 11.93 & \textcolor{bestblue}{\textbf{21.51}} & 4.33 & 4.12 & 6.64 & \textcolor{bestblue}{\textbf{12.58}} & 127.07 & 105.14 & 290.92 & 796.45\\
100 & \textbf{64}  
& 7.65 & 7.09 & 12.05 & 23.02 & 4.31 & 4.08 & 6.78 & 15.41 & 128.65 & 106.24 & 292.17 & 844.65\\
\bottomrule
\end{tabular}
}
\end{center}
\vspace{-0.5cm}
\end{table}

%% file: chapters/tables/ablation_batchsize.tex
\begin{table}[h]
\setlength{\tabcolsep}{2.5pt}
\caption{Varying the batch size. Results are on \textit{AgeDB-DIR} with %$L_1$ loss and 
SQINV re-weighting. We keep $\lambda$ as 2 and $\gamma$ as 100.%, and remove ties in a batch. 
} 
\vspace{-4pt} 
\label{table:hyperparams_bs}
\small
\vskip 0.15in
\begin{center}
\resizebox{0.9\linewidth}{!}{
\begin{tabular}{l|cccc|cccc|cccc}
\toprule[1.5pt]
\textbf{Metrics} &
\multicolumn{4}{c|}{\textbf{MAE~$\downarrow$}}       & \multicolumn{4}{c|}{\textbf{GM~$\downarrow$}}  & \multicolumn{4}{c}{\textbf{MSE~$\downarrow$}}  \\ \midrule
\textbf{Shot} & All   & Many  & Med. & Few   & All   & Many   & Med. & Few & All   & Many  & Med. & Few   \\
 \midrule\midrule

\textsc{Vanilla} %& $/$
&7.77& 6.62& 9.55& 13.67 &5.05& 4.23& 7.01 &10.75& 101.60& 78.40& 138.52& 253.74 \\
\textsc{SQInv} + \textsc{LDS} + \textsc{FDS} %& $/$ %& 1e-3
& {7.55} &  7.01  & {8.24}   & 10.79 & {4.72} & {4.36} & {5.45}  & {6.79} & 99.46 & 84.10 & 112.20 & 209.27\\

\textsc{SQInv} + \textbf{\textsc{RankSim}}, batch size 64  %&
& \textbf{6.91} &  \textbf{6.34} & \textbf{7.79}   & {9.89} 
& \textbf{4.28} & \textbf{3.92} & \textbf{4.88} & {6.89}
& {\textbf{82.10}} & {\textbf{68.60}} & \textbf{102.61} & \textbf{152.84}\\

\textsc{SQInv} + \textbf{\textsc{RankSim}}, batch size 128
& {7.13} & {6.58} & {8.07} & \textbf{9.79}
& {4.40} & {3.95} & {5.38} & {7.02}
& {88.93} & {76.68} & {106.52} & {156.00}\\

\textsc{SQInv} + \textbf{\textsc{RankSim}}, batch size 256
& {7.17} & {6.63} & {7.98} & {10.06}
& {4.53} & {4.18} & {5.24} & \textbf{6.38} & {89.37} & {75.73} & {105.36} & {173.92}\\

\bottomrule
\end{tabular}
}
\end{center}
\vspace{-0.5cm}
\end{table}